\documentclass[11pt]{article}

\usepackage[final]{ACL2023}

\usepackage{times}
\usepackage{latexsym}
\usepackage{tikz}
\usepackage{xcolor}
\usepackage{xcolor}
\usepackage{lipsum} 

\usepackage[utf8]{inputenc}
\usepackage{graphicx}
\usepackage{placeins}

\usepackage{microtype}
\usepackage{listings} 
\usepackage{inconsolata}
\usepackage{import}
\usepackage{microtype}
\usepackage{layout}
\usepackage{tabularx, makecell}
\usepackage{booktabs}
\usepackage{mathrsfs}
\usepackage{amssymb} 
\usepackage{url}
\usepackage{graphicx}
\usepackage{xspace,paralist}
\usepackage{times,latexsym}
\usepackage{amsmath}
\usepackage{appendix}
\usepackage{comment} 
\usepackage{enumitem}
\usepackage{makecell}
\usepackage{multirow}
\usepackage{colortbl}
\usepackage{xcolor}
\usepackage{tablefootnote}
\usepackage{enumitem}  
\usepackage{lipsum}
\usepackage{xspace}
\usepackage{bbding}
\usepackage{pifont}
\usepackage{arydshln}
\usepackage{array}
\usepackage{cleveref}

\definecolor{darkgreen}{rgb}{0.0, 0.5, 0.0}
\definecolor{customblue}{HTML}{0280FB}

\usepackage[russian,english]{babel}

\usepackage{CJKutf8}

\usepackage{float}
\restylefloat{table}

\usepackage[T1]{fontenc}
\usepackage{arydshln}
\usepackage[utf8]{inputenc}

\usepackage{microtype}

\usepackage{inconsolata}
\usepackage{CJK}
\usepackage[T2A,T1]{fontenc}
\AtBeginDocument{%
  \DeclareFontFamilySubstitution{T2A}{\familydefault}{Tempora-TLF}%
}
\title{Instruction Tuning on Public Government and Cultural Data\\ for Low-Resource Language: a Case Study in Kazakh}

\author{
  Nurkhan Laiyk$^1$\thanks{\hspace{0.2cm}These authors contributed equally.}  \quad 
  Daniil Orel$^{1*}$ \quad  
  Rituraj Joshi$^3$ \quad  
  Maiya Goloburda$^1$ \\  
  \textbf{Yuxia Wang}$^1$ \quad  
  \textbf{Preslav Nakov}$^{1,2}$ \quad  
  \textbf{Fajri Koto}$^1$ \\  
  \textsuperscript{1}Mohamed bin Zayed University of Artificial Intelligence \\  
  \textsuperscript{2}Institute of Foundation Models \\  
    \textsuperscript{3}Cerebras Systems \\  
  \texttt{\small \{nurkhan.laiyk,daniil.orel\}@mbzuai.ac.ae} \\
}

\usepackage{graphicx}
\begin{document}
\maketitle

\begin{abstract}


Instruction tuning in low-resource languages remains underexplored due to limited text data, particularly in government and cultural domains. To address this, we introduce and open-source a large-scale (10,600 samples) instruction-following (IFT) dataset,\footnote{\url{https://huggingface.co/datasets/nurkhan5l/kazakh-ift}} covering key institutional and cultural knowledge relevant to Kazakhstan. Our dataset enhances LLMs' understanding of procedural, legal, and structural governance topics.
We employ LLM-assisted data generation, comparing open-weight and closed-weight models for dataset construction, and select GPT-4o as the backbone. Each entity of our dataset undergoes full manual verification to ensure high quality. We also show that fine-tuning Qwen, Falcon, and Gemma on our dataset leads to consistent performance improvements in both multiple-choice and generative tasks, demonstrating the potential of LLM-assisted instruction tuning for low-resource languages.

\end{abstract}

\section{Introduction}

Instruction tuning enhances large language models (LLMs) by fine-tuning them on structured prompts, improving their ability to follow human instructions across various tasks such as question answering and summarization \cite{ouyang2022instructgpt}. While extensive instruction-tuning datasets exist for English, such as FLAN \cite{longpre2023flan}, P3 \cite{sanh2021multitask}, and Dolly \cite{conover2023dolly}, efforts in low-resource languages remain limited. This gap is particularly evident in domain-specific applications where multilingual LLMs often provide generic or inaccurate responses due to a lack of localized training data \cite{li2023bactrian}.

A key challenge in adapting LLMs to underrepresented languages is the scarcity of high-quality instruction data \cite{li2023bactrian}. Multilingual models may process low-resource languages at a technical level \cite{openai2024gpt4o}, but their practical effectiveness is often constrained by an incomplete understanding of region-specific socio-political structures and cultural contexts. For example, when asked about administrative procedures like obtaining a passport in a particular country, models tend to default to well-documented cases rather than providing precise, localized information. Similarly, cultural narratives—such as folklore, literature, and traditions—are often missing from instruction datasets \cite{conover2023dolly}, limiting the models’ ability to generate contextually appropriate responses. While prior work relied on translation \cite{sengupta2023jais} or template-based techniques \cite{cahyawijaya-etal-2024-cendol} to build instruction-tuning datasets, it does not fully reflect the actual local context, as direct translations often fail to capture the nuances of regional governance, customs, and linguistic variations.

Building instruction datasets from scratch is costly, making large-scale manual data collection impractical for many low-resource languages. To address this, we adopt an LLM-assisted dataset generation approach \cite{liu-etal-2022-wanli,cahyawijaya2023instructalign,zhang-etal-2024-llm-assisted}, followed by full human validation. Specifically, we use a single-prompt method where LLMs process high-quality unlabeled text from public government and cultural sources to extract both factual information and corresponding instructions. These domains are highly relevant for real-world applications, but remain underexplored for instruction-tuning, particularly in the context of government data.

\begin{figure*}[h!]
    \centering
    \includegraphics[width=1\textwidth]{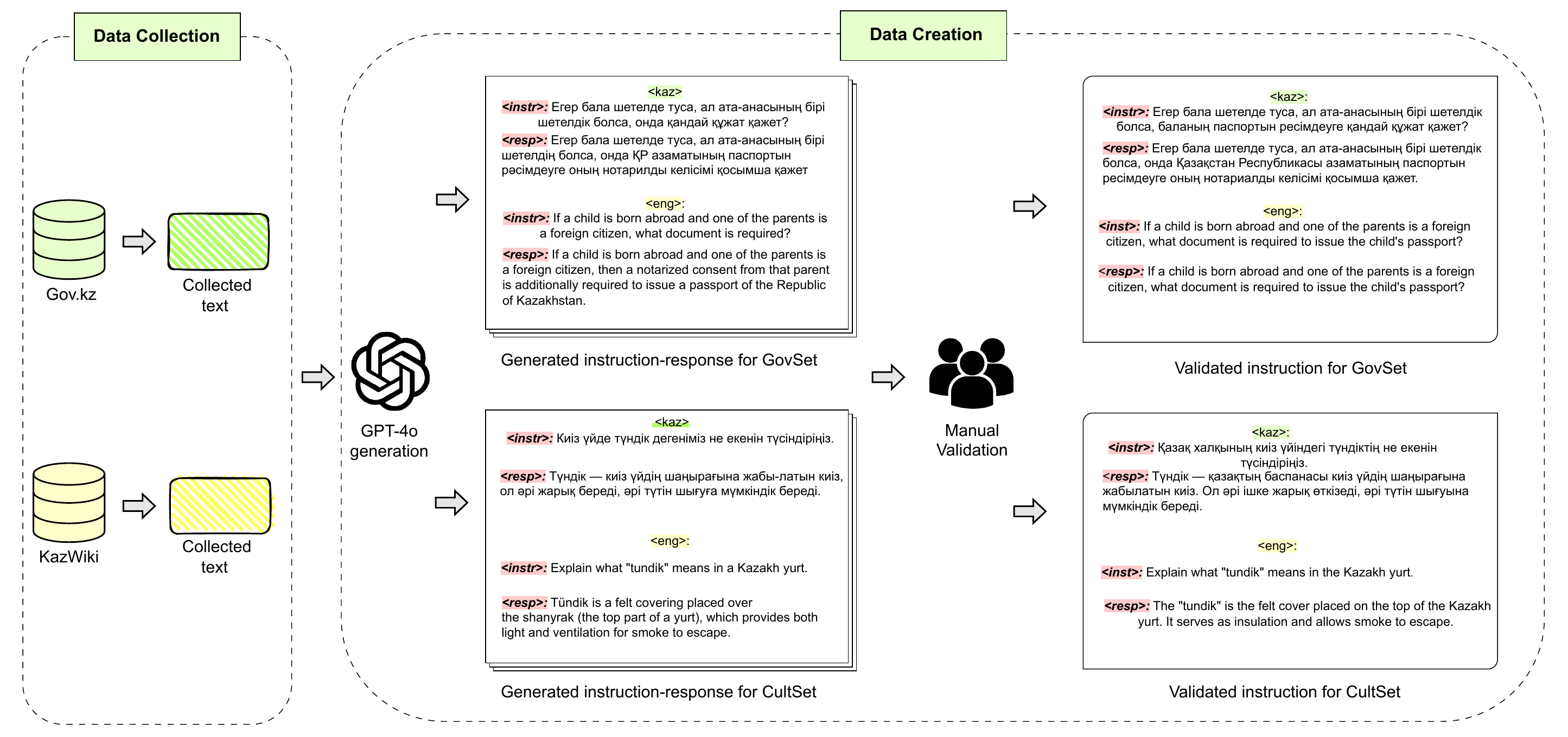} 
    \caption{Overview of the end-to-end process for constructing GovSet and CultSet datasets. English translations are for illustration purposes.}
    \label{fig:pdf-image}
\end{figure*}

To demonstrate the effectiveness of this approach, we introduce an instruction-tuning dataset for Kazakh that integrates both institutional\footnote{Our study incorporates administrative, procedural, legal, structural, and other government-related types of information.} (\texttt{GovSet}) and cultural (\texttt{CultSet}) domains. We choose Kazakh as our case study because it remains underrepresented in NLP \citep{joshi-etal-2020-state}, despite having approximately 20 million speakers. Prior research on Kazakh NLP has primarily focused on classic tasks such as named entity recognition \cite{yeshpanov-etal-2022-kaznerd} and sentiment analysis \cite{yeshpanov2024kazsandra}, leaving more advanced applications like instruction tuning, largely unexplored.  

Our contributions are as follows:
\begin{compactitem}
    
    \item We create an open-source, high-quality, manually verified large scale (10K samples) IFT dataset, which covers both cultural, and institutional knowledge, relevant to Kazakhstan.
    \item We contribute new domain knowledge on essential institutional topics, including procedural, legal, structural, and other key aspects of public governance, enhancing LLMs' understanding of these critical areas.
    \item We compare the efficacy of open-weight and closed-weight LLMs in LLM-assisted dataset construction for low-resource languages and underrepresented cultures.
    \item We demonstrate that fine-tuning on our dataset results in consistent improvements in both multiple-choice and generative tasks. These results highlight the impact of incorporating localized knowledge into instruction tuning and demonstrate the potential of LLM-assisted approaches for expanding instruction datasets in other low-resource languages.
\end{compactitem}

\section{Related Work}
\subsection{Instruction-Tuning Datasets in English}


There are three main strategies for creating English instruction-tuning datasets: human-curated datasets, templatized NLP tasks, and synthetic data generation using LLMs.

Human-curated datasets, such as Open Assistant \cite{kopf2023openassistant} and Dolly \cite{conover2023dolly}, rely heavily on human annotation. While this approach ensures high-quality data, it is expensive and difficult to scale across multiple languages. To reduce costs, datasets like the Public Pool of Prompts (P3) \cite{sanh2021multitask}, SuperNatural Instructions \cite{wang2022super}, and FLAN \cite{longpre2023flan} reformat existing NLP tasks into instruction-based formats. However, these datasets primarily focus on specific NLU tasks rather than general-purpose instruction following, limiting their applicability.

Prior work on instruction dataset creation has largely relied on generating data from existing language models without incorporating external real-world knowledge. Self-Instruct \cite{wang2023self} expands an initial set of human-written instructions by iteratively generating new tasks using the model’s own outputs, while \citet{honovich2022unnatural} create instruction-tuning datasets by conditioning on a few example instructions. These methods rely solely on sampling data from language models or predefined topics, rather than grounding them in external knowledge, making them less suitable for capturing domain-specific or culturally relevant understanding. Unlike these approaches, our work focuses on Kazakh, a low-resource language, and constructs an instruction dataset by leveraging external, factual sources such as governmental and cultural texts, ensuring alignment with real-world contexts.


\subsection{Instruction Tuning Datasets for Medium- to Low-Resource Languages}




While human-curated datasets are often expensive and require native speakers, prior work has explored automatic dataset generation using machine translation for low- to medium-resource languages. \citet{sengupta2023jais} applied this approach to develop JAIS, an Arabic-centric language model, by translating various English instruction datasets into Arabic. Similarly, \citet{li2023bactrian} translated Alpaca \cite{taori2023stanford} into 52 languages. More recently, \citet{CIDAR2024} introduced an Arabic instruction dataset by translating Alpaca into Arabic and then performing manual edits and localization to ensure relevance to the Arabic context. Although this method is scalable, the quality of translation remains inconsistent for low-resource languages, as noted by \citet{li2023bactrian}. Additionally, machine-translated datasets often introduce Anglocentric biases, limiting their ability to capture culturally diverse perspectives.

Several frameworks have been proposed to improve instruction tuning for low-resource languages. MURI \cite{koksal2024muri} generates multilingual instruction datasets using reverse instruction generation and translation, but none of its data have been validated by native speakers. \citet{li-etal-2024-x} improve upon translation-based methods by prompting LLMs in English while requiring responses in low-resource languages (e.g., Urdu), allowing models to leverage their internal knowledge of the target language's local context. However, our approach differs by grounding responses in factual information from reliable sources, such as government data, rather than relying solely on the model’s internal knowledge. Additionally, our dataset undergoes full human validation to ensure accuracy and relevance. Meanwhile, \citet{cahyawijaya2023instructalign} focused on the linguistic aspects of instruction tuning by denoising low-resource language text and prompting models to reconstruct complete sentences. While this enhances fluency, it differs from our approach, which prioritizes grounding instructions in externally verified, domain-specific knowledge rather than refining linguistic quality alone.

\subsection{Existing datasets in Kazakh}
While there has been significant progress in developing Kazakh datasets, the majority of high-quality Kazakh datasets are related to speech~\cite{kaz_asr, mussakhojayeva22_interspeech, mussakhojayeva21_interspeech}. In terms of textual data, existing resources primarily focus on question answering and reading comprehension rather than instruction tuning. For instance, KazQAD \cite{kazqad2024} is a Kazakh open-domain question answering (ODQA) dataset that can be used in both reading comprehension and full ODQA settings, as well as for information retrieval experiments. Similarly, Belebele \cite{bandarkar2023belebele} is another dataset that, while useful for multilingual machine reading comprehension, is not explicitly designed for instruction tuning. Belebele covers 122 languages, including Kazakh, and comprises 900 multiple-choice questions associated with 488 distinct passages from the Flores-200 dataset.



Despite progress in Kazakh NLP resources, no existing instruction-tuning dataset incorporates cultural or domain-specific knowledge, focuses on real-world applications, and undergoes full human validation. This limits the ability of LLMs to process Kazakh-language instructions effectively in practical and locally relevant contexts, which we aim to address in this work.


\section{Background} 
\textbf{Kazakh Cultural Heritage.} Kazakhstan has a rich cultural heritage that reflects a blend of nomadic traditions, Soviet influences, and modern developments. The country's nomadic heritage is evident in many aspects of daily life, from its architecture, with \textit{yurts} (traditional felt tents) still used in rural areas, to its customs of communal gatherings and feasts, such as the celebration of the Turkic New Year, \textit{Nauryz}.


Kazakhstan's Soviet past has also left a lasting imprint on its culture. Many cities still bear the architectural marks of Soviet planning, while the era also shaped the country's education and scientific institutions, fostering a strong tradition in mathematics and engineering---most notably reflected in the \textit{Bai\-ko\-nur Cos\-mo\-drome}, the world's first and largest space launch facility.

Alongside all of this, modern developments have transformed Kazakhstan. The capital, Astana, is a prime example of this shift, with its futuristic skyline and ambitious urban projects. Investments in technology, renewable energy, and digital infrastructure have propelled Kazakhstan onto the global stage, while cultural revitalization efforts have fostered a renewed interest in the Kazakh language, music, and art.

At its core, Kazakhstan's culture is shaped by beliefs and values, social practices, language, artistic expression, and material culture. Understanding these components is crucial for ensuring accurate and meaningful representation.

\paragraph{Kazakhstan's Institutional Structure and Public Governance.}

Kazakhstan is a presidential republic that has prioritized modernization since its independence in 1991, particularly in governance and legal systems. The 1995 Constitution established the legal foundation, defining citizens’ rights and the structure of government. A major step in this modernization has been the digitalization of public services. Kazakhstan ranks among the top 25 countries in the UN E-Government Development Index (EGDI)~\citep{UN_EGDI}, with the \textit{eGov} platform serving as a centralized portal for services like business registration, tax payments, and social benefits.

These efforts reflect a broader national context shaped by Kazakhstan’s cultural heritage, nomadic traditions, and growing digital infrastructure. Platforms like \textit{eGov} highlight the integration of technology into daily governance. As the country continues to modernize, it is essential that language models accurately represent these unique characteristics to support cultural understanding and global relevance.

\section{Data}
\subsection{Document Source}

\paragraph{\texttt{GovSet}} We manually collected 1,376 texts from the official Kazakhstan e-Government portal (gov.kz\footnote{\url{https://www.gov.kz}}), the primary and most comprehensive platform for all public services, governmental processes, and administrative resources in the country. As the central hub for Kazakhstan's digital governance, \texttt{gov.kz} consolidates a wide range of essential information into a single system, covering diverse aspects of public administration, legal frameworks, citizen services, and governmental initiatives. 
By incorporating these texts, we ensure that the dataset captures essential institutional aspects of life in Kazakhstan, including its governmental structure and public services. This enrichment enhances instruction-tuning applications, making them more linguistically appropriate and contextually informed.

\paragraph{\texttt{CultSet}} We automatically collected 4,400 texts from Kazakh Wikipedia,\footnote{\href{https:_/_/kk.wikipedia.org}{kk.wikipedia.org}} specifically focusing on pages related to Kazakh culture. These pages were identified based on metadata that explicitly indicated their relevance to Kazakh cultural topics. 
The parsed texts include various aspects of Kazakh traditions, heritage, arts, and historical practices, providing a rich source of culturally relevant content.
This ensures that the dataset reflects the depth and diversity of Kazakh culture, making it suitable for instruction-tuning tasks that require a culturally grounded perspective.


\subsection{LLM-assisted Data Generation}

We benchmark one open-weight LLM: LLaMA 3.1-70B \cite{touvron2023llama}, and three closed-weight LLMs: GPT-4o \cite{openai2024gpt4o}, Gemini-1.5 \cite{google2024gemini15}, and Claude-3.5-Sonnet \cite{anthropic2024claude}, to assess their effectiveness in assisting dataset creation. These models were selected based on their strong performance in multilingual benchmarks. However, their capability in generating instruction datasets specific to Kazakh government and cultural data remains uncertain.


We design a prompt (see Appendix~\ref{sec:prompts}) that instructs LLMs to first extract factual information from a given Kazakh document and then generate an instruction dataset based on the extracted content. Table~\ref{tab:ds_stats} provides detailed statistics on the source documents and the resulting instruction fine-tuning (IFT) dataset using GPT-4o. Specifically, we use 4,400 Kazakh cultural Wikipedia documents and 1,376 Kazakh government data sources, generating a total of 10,600 IFT instances. Of these, 58\% belong to the government public data category (\texttt{GovSet}), while the remaining samples are derived from Wikipedia (\texttt{CultSet}). Examples of generated IFT data can be found in Table~\ref{tab:example_govset} and Table~\ref{tab:example_cultset}.

\paragraph{Human Evaluation Across LLMs}  
For each LLM, we sampled 100 generated IFT instances, drawn from 25 randomly selected \texttt{GovSet} and 25 \texttt{CultSet} documents. Additionally, we randomly sampled 100 instances from MURI~\cite{koksal2024muri}, which also includes Kazakh IFT data, to provide a comparative quality assessment. Two native Kazakh speakers were recruited to manually evaluate the generated data based on the following criteria:
\begin{compactitem}
    \item \textbf{Correctness}: The factual accuracy and alignment with the original text. A high score indicates that the generated pair adheres closely to the source material without introducing errors or inaccuracies.
    \item \textbf{Fluency}: The grammatical and stylistic quality of the generated text. A higher score reflects well-structured, natural, and polished language.
    \item \textbf{Completeness}: The degree to which the instruction-response pair is clear, contextually grounded, and free from ambiguity. High scores indicate that the pair is fully self-contained, with enough context to make it understandable. 
\end{compactitem}
All criteria were rated on a Likert scale from 1 to 5, with 5 representing the highest quality. A detailed evaluation rubric is provided in Table~\ref{tab:multilingual-issues}.

Table~\ref{tab:llm-instruction} presents the quality assessment of various LLMs in generating IFT data for Kazakh. The inter-annotator agreement, measured using Pearson correlation, is high (ranging from 0.68 to 0.70) across correctness, completeness, and fluency, indicating strong reliability in the evaluation process (see Appendix \ref{app:inner-annot-prelim} for further details).

Among the evaluated models, GPT-4o achieved the highest performance across all three criteria. In contrast, LLaMA-3.1 (70B) lagged significantly, scoring nearly 0.8–1 point lower in all aspects. Notably, MURI's quality was lower than GPT-4o despite both relying on OpenAI models. This discrepancy is likely due to MURI’s reliance on machine translation, where Kazakh text is first translated into English before generating instructions, followed by a final back-translation into Kazakh. This multi-step translation process can introduce errors due to cumulative translation inaccuracies. Additionally, MURI is entirely LLM-generated without human validation, further affecting its quality.

\begin{table}[t]
\centering
 \resizebox{0.9\linewidth}{!}{
    \begin{tabular}{lccc}
    \toprule
    \textbf{Model} & \textbf{Correctness} & \textbf{Completeness} & \textbf{Fluency} \\ 
    \midrule
    Llama 3.1 (70B)  & 3.54 & 3.45 & 3.07 \\
    Claude & 3.74 & 3.48 & 3.09 \\
    Gemini 1.5  & 3.85 & 3.64 & 3.32 \\
    GPT-4o & \textbf{4.38} & {\textbf{4.29}} & \textbf{4.04}  \\ 
    \hdashline
    MURI & 3.87 & 3.52 & 3.41\\
    \bottomrule
    \end{tabular} 
} 
\caption{Human evaluation on LLM-generated instruction datasets.}

\label{tab:llm-instruction}
\end{table}



\begin{table}[t]
\centering
\renewcommand{\arraystretch}{1.2} 
 \resizebox{0.9\linewidth}{!}{
\begin{tabular}{lrr}
\toprule
 & \textbf{CultSet} & \textbf{GovSet} \\ \midrule
Collected text & 4,400 & 1,376 \\ 
Avg. lengths (\#char) of collected text & 245 & 179\\
\cdashline{1-3} 

Generated IFT pairs & 4,400 & 6,200 \\
Avg. lengths (\#char) of instruction & 85 & 76\\
Avg. length (\#) of output & 453 & 215\\
\# of unique tokens & 62,449 & 24,304\\
\bottomrule
\end{tabular}}
\caption{Statistics of GPT-4o generated IFT dataset.}
\label{tab:ds_stats}
\end{table}

\begin{table}[t]
\small
\centering
\resizebox{0.65\linewidth}{!}{
\renewcommand{\arraystretch}{1.2}
\begin{tabular}{lcc}
    \toprule
    \multirow{2}{*}{\textbf{Error Type}} & \multicolumn{2}{c}{\textbf{\% of Questions}} \\
    \cmidrule(lr){2-3}
    & \textbf{CultSet} & \textbf{GovSet} \\
    \midrule
    No error & 28.32\% & 19.47\% \\
    \hdashline
    Wrong language & 0.07\% & 0.14\% \\
    \textbf{Structural} & \textbf{28.45\%} & \textbf{33.58\%} \\
    Grammatical & 25.24\% & 28.73\% \\
    Lexical & 17.92\% & 18.08\% \\
    \bottomrule
\end{tabular}}
\caption{Distribution of error types in GPT-4o-generated IFT data from \texttt{CultSet} and \texttt{GovSet}, identified during manual post-editing.}
\label{tab:error_analysis}

\end{table}

\subsection{Manual Post-Editing}

Given GPT-4o’s strong performance, we use it for large-scale IFT data generation while ensuring quality through full human verification. We employ 12 expert annotators, all native Kazakh speakers with advanced degrees in World Languages, Literature, or Political Science from top Kazakhstani universities. Their extensive experience—having lived in Kazakhstan for over 25 years—equips them with the necessary linguistic and cultural expertise.

To maintain consistency, annotators received detailed guidelines outlining task objectives, evaluation criteria, and examples of high-quality IFT pairs (see Appendix \ref{app:annot-guide-appendix}). They were responsible for manually reviewing and correcting errors in the generated data. Before starting the main annotation process, all candidates completed a pilot task to assess their understanding of project requirements and their ability to refine IFT pairs accurately. Only those who met the evaluation criteria were selected. Each annotator's workload was equivalent to five full working days, and they were compensated fairly based on Kazakhstan’s monthly minimum wage. To accommodate flexibility, annotators were given up to one month to complete the task while working part-time.

Table~\ref{tab:error_analysis} summarizes the error types identified during manual post-editing of GPT-4o-generated data across the two document sources. Annotators found that \texttt{CultSet} had a higher proportion of "No error" cases (28.32\%) compared to \texttt{GovSet} (19.47\%), suggesting variations in data quality.

Structural errors were the most common in both datasets, accounting for over 28\% in \texttt{CultSet} and 33\% in \texttt{GovSet}. These errors involve grammatically correct but poorly structured responses, including issues with logical flow, organization, and unnatural phrasing for a Kazakh speaker. Additionally, grammatical and lexical errors were frequently observed, with annotators noting that GPT-4o occasionally replaces Kazakh words with Russian equivalents, even when the correct Kazakh term is explicitly provided in the original text. For a detailed breakdown of annotator observations, see Appendix~\ref{app:annot-comments}.

\subsection{Final Data Overview}

As shown in Table~\ref{tab:ds_stats}, the final dataset consists of 4,400 \texttt{CultSet} and 6,200 \texttt{GovSet} IFT instances, totaling 10,600 high-quality samples. We split the dataset into 90\% training and 10\% test, where the training data is used for full fine-tuning of LLMs, and the test set is used for generation evaluation in our experiments.

\begin{figure*}[h!]
    \centering
    \begin{minipage}[b]{0.45\textwidth}
        \includegraphics[scale=0.35]{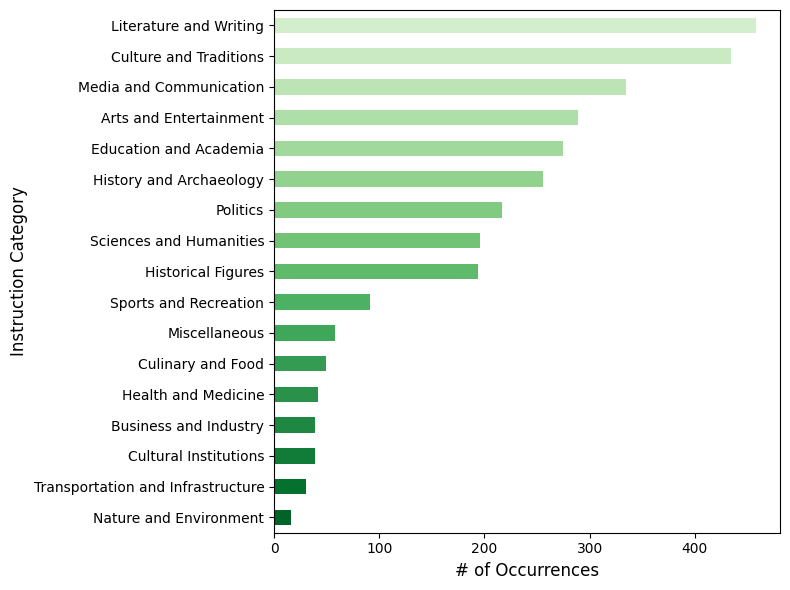}
        \caption*{(a) CultSet}
    \end{minipage}
    \hfill
    \begin{minipage}[b]{0.45\textwidth}
        \centering
        \includegraphics[scale=0.35]{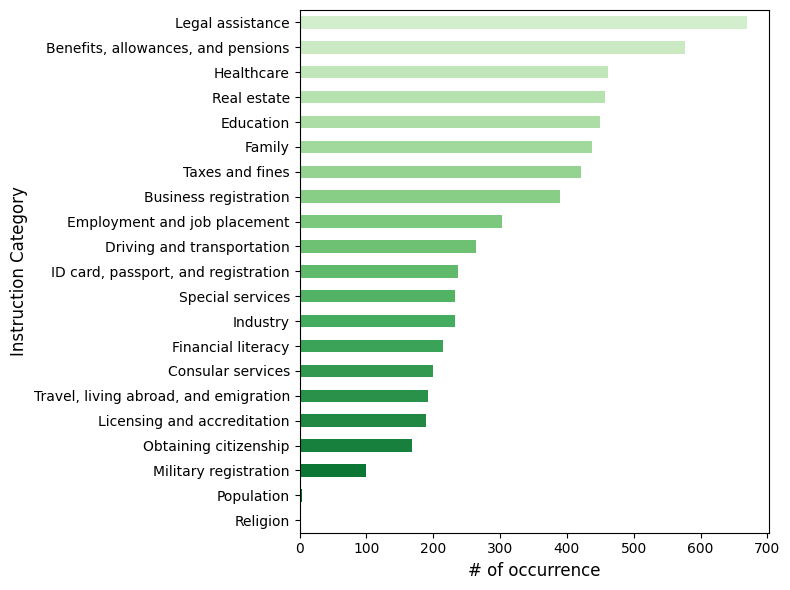}
        \caption*{(b) GovSet}
    \end{minipage}
    \caption{Topic distribution of GPT-4 generated IFT dataset in \texttt{CultSet} and \texttt{GovSet}.}
    \label{fig:distribution-all}
\end{figure*}

Since both \texttt{CultSet} and \texttt{GovSet} are topic-based, we include their respective topics as metadata in the final IFT dataset (see Table~\ref{tab:category_exp_gov} and Table~\ref{tab:category_exp_culture} for topic definitions). Figure~\ref{fig:distribution-all} illustrates the topic distribution of the dataset. The most common topics in \texttt{CultSet} include Kazakh literature, traditions, and media, while \texttt{GovSet} primarily covers legal assistance, the healthcare system, real estate laws, and education in Kazakhstan. Examples of GPT-4o-generated IFT data can be found in Table~\ref{tab:example_govset} and Table~\ref{tab:example_cultset}.

Table~\ref{tab:ds_stats} further highlights a notable difference between the two subsets: the average output length in \texttt{CultSet} is significantly longer and includes more unique tokens than \texttt{GovSet}. This difference stems from the nature of \texttt{GovSet} responses, which are strictly factual and concise, whereas \texttt{CultSet} responses tend to be more diverse and expressive.

\begin{figure*}[ht!]
    \centering
    \begin{minipage}[b]{0.45\textwidth}
        \centering
        \includegraphics[scale=0.4]{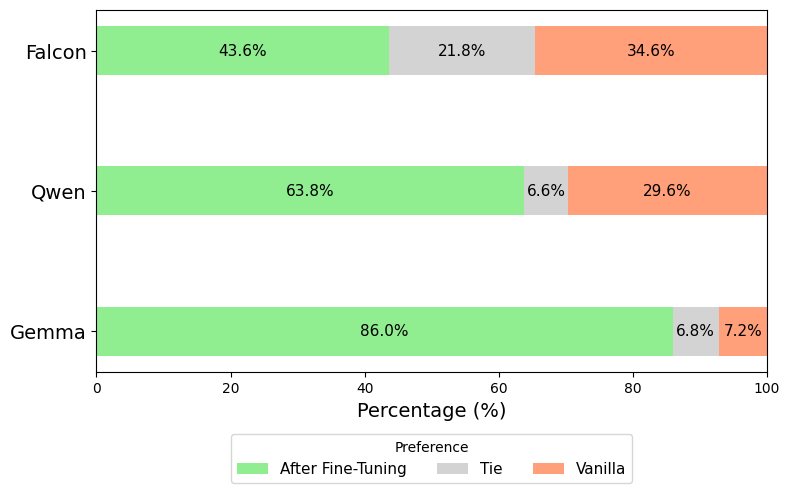}
        \caption*{(a) CultSet}
    \end{minipage}
    \hfill
    \begin{minipage}[b]{0.45\textwidth}
        \centering
        \includegraphics[scale=0.4]{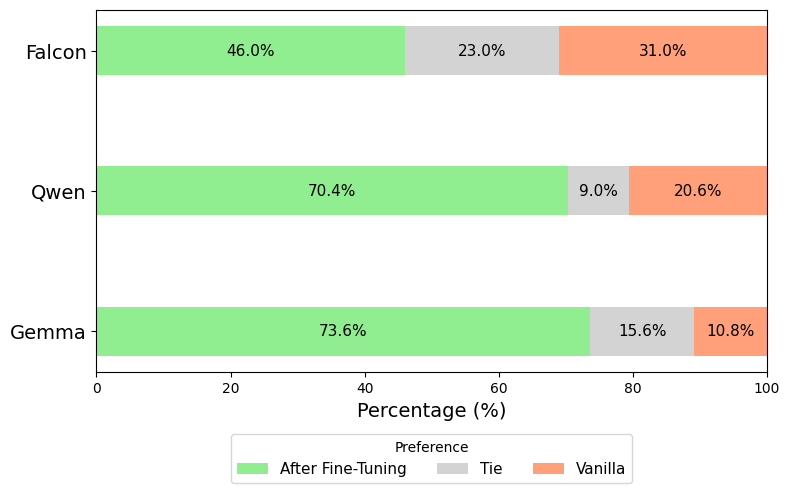}

        \caption*{(b) GovSet}
    \end{minipage}
    \caption{Distribution of preferences for (a) \texttt{CultSet} and (b) \texttt{GovSet} datasets across models. The charts illustrate the percentage of 'Tie', 'Vanilla', and 'After Fine-Tuning' preferences in each dataset.}
    \label{fig:human_fg_1000}
\end{figure*}
\section{Experiments}

We conducted two experiments: multiple-choice questions (MCQ) and text generation evaluation. We will detail each evaluation in the following sections.

\paragraph{Model Selection} For both MCQ and generation evaluations, we use three instruction-tuned models: Gemma-2-9b-instruct (Gemma) \cite{gemma2}, Qwen-2.5-7b-instruct (Qwen) \cite{qwen25}, and Falcon-3-10b-instruct (Falcon)~\cite{Falcon3}. While these LLMs offer multilingual capabilities, none were specifically trained for Kazakh, allowing us to assume that our IFT data is novel to them.

\paragraph{Fine-tuning} We performed full fine-tuning on Gemma-2-9b-instruct (Gemma), Qwen-2.5-7b-instruct (Qwen), and Falcon-3-10b-instruct (Falcon) using the AdamW optimizer with hyperparameters $\beta_1 = 0.9$, $\beta_2 = 0.95$, $\epsilon = 1e{-5}$, and a weight decay of 0.1. We scaled the gradient norms using a maximum norm clipping value of 1.0. The learning rate was kept constant throughout the fine-tuning without any warm-up or decay with a value of $1e{-6}$ for Gemma and Falcon, and $1e{-5}$ for Qwen. The batch size used was 16, and we packed multiple documents until the maximum sequence length was 8,192 tokens. Cross-document attention is disabled by modifying attention masks so the tokens of a document only attend to the tokens from the same document in a causal way. No adjustments were made to the original tokenizer for each model.

\paragraph{Baseline} As a baseline, we include the Kazakh Alpaca dataset,\footnote{\url{https://huggingface.co/datasets/AmanMussa/kazakh-instruction-v2}} which has been translated and localized into Kazakh. For each model, we conduct full fine-tuning with (1) our training dataset, (2) Alpaca, and (3) a combination of Alpaca and our training dataset.

\begin{table}[t!]
\centering

\renewcommand{\arraystretch}{1.2} 
\resizebox{0.9\columnwidth}{!}{ 
\begin{tabular}{lccccc}
\toprule
\textbf{Model} & \textbf{Vanilla} & \textbf{RAG} & \textbf{Alpaca} & \textbf{Ours} & \textbf{Alpaca + Ours}  \\  
\midrule
\multicolumn{6}{l}{\cellcolor{blue!7}\textbf{Dastur}} \\
Gemma & 0.498 & {0.533} & 0.513 & 0.543 & \textbf{0.566}  \\

Qwen & 0.403 & {0.410} & 0.421 & 0.443 &\textbf{0.465 }  \\

Falcon & 0.450 & {0.460} & 0.458 & 0.464 & \textbf{0.471 }   \\
\midrule
\multicolumn{6}{l}{\cellcolor{blue!7}\textbf{Constitution}} \\
Gemma & 0.600 & {\textbf{0.655}} & 0.627 & 0.640 & {0.650}  \\

Qwen & 0.520 & {0.523} & 0.609 & 0.670 & \textbf{0.680}  \\

Falcon & 0.430 & {0.386} & 0.450 & 0.490 & \textbf{0.520}  \\
\midrule
\multicolumn{6}{l}{\cellcolor{blue!7}\textbf{Human Rights and Society}} \\
Gemma & 0.405 & 0.450 & 0.430 & 0.465 & \textbf{0.480}  \\

Qwen & 0.300 & 0.325 & 0.330 & 0.365 & \textbf{0.375}  \\

Falcon & 0.215 & 0.220 & 0.234 & 0.250 & \textbf{0.275}  \\
\bottomrule
\end{tabular}}
\caption{Zero-shot accuracies of language models in different datasets: (1) Dastur, (2) Constitution, and (3) Human Rights and Society}
\label{tab:result_mcq}
\end{table}









\subsection{Multiple-choice Question Evaluation}
\label{sec:mcq}

\paragraph{Dataset}
A dedicated open-source Kazakh NLP community\footnote{\url{https://huggingface.co/kz-transformers}} has collaboratively developed and crowd-sourced multiple hand-crafted benchmarks to assess the factual knowledge of LLMs in Kazakh.  We use three multiple-choice question (MCQ) datasets: (1) Dastur-MC~\cite{horde_dastur_kk2024}, which evaluates knowledge of Kazakh traditions, (2) Kazakh Constitution-MC~\cite{horde_constitution_kk2024}, which focuses on Kazakhstan’s legal system, and (3) Kazakh Unified National~\cite{horde_unt_kk2024}, which assesses citizen's rights, legal protections, and societal knowledge (referred to as the "Human Rights and Society" dataset).\footnote{Examples of test questions are provided in Appendix \ref{sec:mcq_samples}.}

Each dataset consists of multiple-choice questions with four answer options, only one of which is correct. We selected these evaluation benchmarks because they align with the focus of our instruction fine-tuning dataset and are not derived from our document sources (\texttt{CultSet} and \texttt{GovSet}). These datasets cover culturally significant topics, legal frameworks, and citizen-government interactions, reflecting real-world applications that our fine-tuned models aim to support.

Since no documented quality assurance process was available for the three datasets, we conducted a manual verification to ensure the accuracy of the questions. To maintain a fair and valid comparison, only the manually verified samples were used in our evaluation. For the Dastur-MC dataset, we randomly sampled 300 questions and manually verified their correctness. The same process was applied to the Kazakh Constitution-MC and Human Rights and Society datasets, with 200 randomly selected questions from each.

\paragraph{Setup}
In addition to the fine-tuned models, we include retrieval-augmented generation (RAG) without fine-tuning to estimate the upper bound of the original models' performance. For RAG, we use BM25 encoding, as no specialized Kazakh retrieval encoder is available. For each question, we retrieve the top two matching text chunks (each 256 symbols long) from the training texts of our IFT corpus and provide them as additional context.

To assess the capabilities of the model, we use the \texttt{LM Eval Harness}~\cite{eval-harness} framework in a zero-shot setting. During evaluation, the answer is selected based on the alphabetical option with the highest likelihood.


\paragraph{Result}
Table~\ref{tab:result_mcq} presents the zero-shot evaluation results across different models and techniques. Overall, our fine-tuned dataset consistently outperforms other approaches across datasets and models. The only exception is the Constitution dataset, where RAG performs better with Gemma. Models fine-tuned on Kazakh Alpaca show some improvement, though it remains lower than that achieved with our instruction fine-tuning (IFT) dataset.

Combining parts of our IFT dataset with the translated Alpaca dataset yields the highest performance gains. This aligns with prior studies~\cite{mixingup,demystifying}, which suggest that incorporating general chat instructions alongside domain-specific ones enhances model performance.

For RAG-enhanced models, performance generally exceeds that of the vanilla models, except for Falcon on the Constitution dataset. However, fine-tuned models consistently achieve higher scores than their RAG-enhanced counterparts. We hypothesize that this is due to the models' limited proficiency in Kazakh, which may hinder their ability to fully understand the retrieved context. As a result, despite the additional information provided by RAG, the models may struggle to extract the necessary details to select the correct answer in MCQs.

\begin{table}[t]
    \centering
    \resizebox{0.8\linewidth}{!}{
    \begin{tabular}{lcccc}
        \toprule
        & \multicolumn{2}{c}{\textbf{CultSet}} & \multicolumn{2}{c}{\textbf{GovSet}} \\
        \cmidrule(lr){2-3} \cmidrule(lr){4-5}
        & \textbf{Vanilla} & \textbf{After FT} & \textbf{Vanilla} & \textbf{After FT} \\
        \midrule
        Gemma  & 15.76 & \textbf{24.87} & 16.12 & \textbf{25.10} \\
        Falcon & 25.96 & \textbf{27.98} & 26.17 & \textbf{28.70} \\
        Qwen   & \textbf{27.64} & 26.63 & \textbf{30.27} & 28.42 \\
        \bottomrule
    \end{tabular}}
    \caption{ROUGE-L comparison on \texttt{CultSet} and \texttt{GovSet} before and after fine-tuning.}
    \label{tab:rouge}
    
\end{table}

\begin{table*}[ht]
    \centering
    \resizebox{0.7\linewidth}{!}{
    \begin{tabular}{llccc ccc}
        \toprule
        & & \multicolumn{3}{c}{\textbf{CultSet}} & \multicolumn{3}{c}{\textbf{GovSet}} \\
        \cmidrule(lr){3-5} \cmidrule(lr){6-8}
        & & \textbf{Precision} & \textbf{Recall} & \textbf{F1} & \textbf{Precision} & \textbf{Recall} & \textbf{F1} \\
        \midrule
        \multirow{3}{*}{\textbf{Vanilla}} & Gemma  & 29.26 & 33.47 & 30.92 & 27.36 & 34.81 & 30.39 \\
                                                            & Falcon & 23.29 & 28.17 & 25.20 & 20.38 & 24.68 & 22.11 \\
                                                            & Qwen   & 40.58 & 47.46 & 43.40 & 36.57 & 44.14 & 39.50  \\
     
        \midrule
           \multirow{3}{*}{\textbf{ After Fine-Tuning}} & Gemma  & 41.94 & 46.36 & 43.62 & 40.27 & 44.90 & 42.00 \\
                                                    & Falcon & 24.59 & 29.68 & 26.64 & 23.78 & 27.73 & 25.36 \\
                                                    & Qwen   & 39.64 & 45.40 & 41.82 & 36.28 & 40.20 & 37.59 \\
        \bottomrule
    \end{tabular}}
    \caption{BERTScore Precision, Recall, and F1 for \texttt{CultSet} and \texttt{GovSet}.}
    \label{tab:bertscore}
\end{table*}

\subsection{Generation Evaluation}
\label{sec:rogue}

We evaluate generation performance using our test set, which consists of 500 questions from both \texttt{CultSet} and \texttt{GovSet} (excluded from fine-tuning). We compare the best-performing models from Section~\ref{sec:mcq} against their vanilla counterparts. In this section, "After Fine-Tuning" refers to models fine-tuned on Alpaca + Our Data, while "Vanilla" refers to the original instruct models.

\paragraph{Automatic Evaluation with ROUGE and BERTScore} As shown in Table~\ref{tab:rouge}, fine-tuned models generally outperform their vanilla counterparts, except for Qwen, where fine-tuning results in a lower ROUGE-L score \cite{lin-2004-rouge}. However, a lower ROUGE-L does not necessarily indicate worse performance—it may be due to Qwen generating different phrasings compared to the gold answers.

To further validate the quality of generated responses, we also evaluate BERTScore \cite{zhangbertscore}. We use Kaz-RoBERTa \cite{Sagyndyk2025KazRoBERTaConversational}
as the encoder model, as it is one of the few open-source Kazakh-language transformers. The BERTScore results in Table~\ref{tab:bertscore} align well with the ROUGE-L scores. However, since Kazakh is a low-resource language, BERTScore should be considered a reference point rather than a definitive metric, as Kaz-RoBERTa embeddings may not perfectly capture synonym relationships.

\paragraph{Preference Evaluation with GPT-4o}
We conducted a 1-to-1 preference evaluation using the LLM-as-a-judge approach. Specifically, we prompted GPT-4o to compare responses from different models and determine whether each response wins, loses, or ties. The prompt includes the instruction and the gold response as context for GPT-4o.\footnote{The prompt used for comparison is provided in Appendix~\ref{sec:pref-eval}.} As shown in Figure~\ref{fig:human_fg_1000}, the results align with ROUGE-L and BERTScore, confirming that fine-tuned models generally produce improved outputs. Compared to Falcon, Qwen and Gemma exhibit more significant improvements (63\%–80\% winning rate), likely because their pre-trained versions were less optimized for the task, making fine-tuning more impactful.

Additionally, we analyze the win rate across topics in \texttt{CultSet} and \texttt{GovSet}, as shown in Appendix~\ref{app:preference-category-eval-results}. The results indicate that the impact of fine-tuning varies by topic and is not always consistent. In \texttt{CultSet}, fine-tuning Qwen with our IFT data yields the most improvement in Cultural Institutions and Culture \& Traditions, while the gains are smaller in Science \& Humanities and even lead to a decline in performance for Education \& Academia. In \texttt{GovSet}, fine-tuning Qwen with our dataset significantly enhances performance in Legal Assistance, though the improvement is less noticeable in Employment-related topics.


\begin{figure}[t!]
    \centering
    \includegraphics[width=\linewidth]{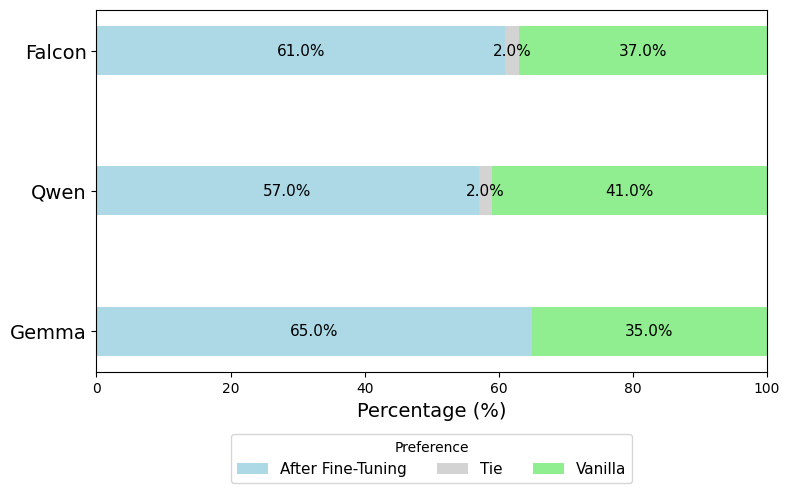}
    \caption{Conversational data preference evaluation.}
    \label{fig:conversational_preferences}
    \vspace{-0.5cm}
\end{figure}

While LLM-based evaluations provide scalable comparisons, they may not fully capture human judgment nuances, making human evaluation essential for validating model preferences. Therefore, three human annotators conducted a preference evaluation on a randomly sampled 100 examples for each model (Gemma, Qwen, and Falcon) across both \texttt{CultSet} and \texttt{GovSet}. Their judgments were compared against the GPT-based preference evaluation \textbf{to assess alignment}. We computed Cohen’s Kappa between GPT-4o and the annotators, obtaining 0.63 for \texttt{CultSet} and 0.68 for \texttt{GovSet}, indicating substantial agreement. We have also calculated the agreement rate between annotators (detailed in Appendix \ref{app:inner-annot-gen-eval}). The results show that GPT's alignment with human preferences is moderate, with better agreement on \texttt{GovSet} than \texttt{CultSet}.

\paragraph{Conversational Evaluation.} As an extension of these experiments, we generated a set of 100 conversations for both \texttt{CultSet} and \texttt{GovSet} combined, covering topics presented in Figure \ref{fig:distribution-all}. These conversations were intentionally left unfinished using a special prompt, as detailed in Appendix \ref{sec:pref-eval}. Both the original and fine-tuned models were tasked with generating the most appropriate continuation for each conversation. Examples of the resulting texts are shown in Appendix~\ref{sec:conversational-data-sample}.
To evaluate the quality of the responses, we employed an LLM-as-a-judge framework. The results, presented in Figure \ref{fig:conversational_preferences}, indicate that models fine-tuned on domain-specific data produced significantly more coherent and contextually appropriate responses compared to their pre-fine-tuning counterparts.
We also see that in the conversational settings there are less ties, compared to simple question answering.





\section{Conclusion and Future Work}
We introduced a culturally and institutionally aligned instruction-tuning dataset for Kazakh, 
aiming to enhance practical knowledge representation and address the specific needs of public governmental data processing in Kazakh.
Through a carefully designed data collection pipeline,
we generated instruction-tuning examples using GPT-4o and ensured their quality via a manual correction and localization to capture Kazakh linguistic and cultural nuances accurately.

The evaluation results show that this approach substantially improved the model's factual knowledge and the understanding of low-resource languages.
It also shows that after such fine-tuning, the model's responses are much better in terms of correctness and soundness, as assessed by native speakers and LLM as a judge.

In future work, we plan to apply this methodology to other languages and dialects. We further aim to work towards streamlining and automating the process as much as possible. 
We will also focus more on the modeling part of the experiments, and open-source culturally and institutionally relevant models for low-resource languages, including Kazakh.

\section{Limitations}
We aim to establish a robust instruction-tuning dataset for Kazakh, authentically reflecting the cultural and linguistic richness of the language. Unlike many existing datasets, which rely on translated resources or machine-generated responses, our dataset is entirely crafted from Kazakh-specific content, ensuring greater alignment with the cultural values and linguistic nuances of the region. However, we recognize several limitations in our work:
\begin{compactitem}
\item \textbf{Cultural Representation:} The dataset emphasizes topics deeply rooted in Kazakh culture, traditions, and societal norms, ensuring relevance and cultural authenticity. However, certain culturally sensitive topics, such as those involving religious matters, were intentionally omitted to avoid controversy and maintain neutrality.
\item  \textbf{Language Variations:} Kazakh is a rich language with significant regional variations in vocabulary and usage. While our dataset primarily focuses on standard Kazakh, it does not explicitly account for regional dialects or variations, potentially limiting its applicability to speakers outside the standard dialect's scope.
\item \textbf{Modeling Limitations:} Our works is a proof of concept, and it was not aimed at creation of SOTA models for Kazakh. That is why we experiment with smaller models and do not apply any training tricks such as tokenizer adaptaion for Kazakh.
\item \textbf{Possible Data Drift:} We also acknowledge that despite of being very conservative by nature, some institutional procedures can change over time, that is why it is possible that the data provided in our IFT dataset will get less actual. To handle this issue we are planning updating the datasets annually.
\end{compactitem}

\section{Ethics}
We adhered to the internal policies of web resources while scraping data and included only publicly available information verified by authorities.

While our method enhances LLMs' understanding of Kazakhstan's institutional nuances, users should not blindly trust generated responses. LLM outputs serve as a starting point, and users remain responsible for fact-checking due to potential hallucinations.

All human subjects in our study provided informed consent, were fully aware of the study's objectives, and had the right to withdraw at any time. They were also appropriately compensated as part of their job.








\bibliography{acl2023}
\bibliographystyle{acl_natbib}
\newpage
\appendix

\section{Prompts Details}
\subsection{Prompt  for preference evaluation}
\label{sec:pref-eval}
\begin{lstlisting}[basicstyle=\ttfamily\scriptsize, breaklines=true]
You are given 2 responses, and a Golden Label. Please, decide which of the responses is the best (you have to take into account its factual correctness, and readability)

    Response 1: {pred1}
    Response 2: {pred2}
    Golden Label: {gold_label}

Return 1 if the first response is better, 2 if the second one is better, 0 if they are equally good. Return only the number
\end{lstlisting} 

\subsection{Prompt for creating conversational data }
\label{sec:pref-eval}
\begin{lstlisting}[basicstyle=\ttfamily\scriptsize, breaklines=true]
Prompt: 
Instruction: {input_instruction}
Output: {output}
You are creating conversational data between two people discussing {topic} in Kazakhstan. The conversation should:  

- Start with a general but relevant topic that smoothly leads into the instruction's question.  
- Be short and concise, where Person 1 initiates the discussion and then asks a question with a meaning similar to the instruction at the end.  
- Person 2 should respond but not reveal the output
- the conversation ends with "Person 2 says: ...".  
- Maintain natural, formal dialogue relevant to government regulations in Kazakhstan.  

Example Format:  
Person 1 says: [General opening statement leading to the topic]  
Person 2 says: [Relevant response that naturally progresses the discussion]  
Person 1 says: [Rephrased instruction question]  
Person 2 says: ...
\end{lstlisting}

\subsection{Prompts for Generation Instruction Dataset}
\label{sec:prompts}

\begin{lstlisting}[basicstyle=\ttfamily\scriptsize, breaklines=true, aboveskip=0pt, belowskip=0pt, lineskip=-2pt, xleftmargin=0pt]
You are given a text: {TEXT}.
I want to extract all the facts from the given text. Based on the extracted facts, I want you to create instruction fine-tuning pairs in Kazakh. 
The pairs may fall into the following categories, but you are free to use other relevant categories if appropriate:
- Is it true that ...
- Explain ...
- Describe ...
- List the steps ...
For each category, provide a clear instruction.
The instructions MUST incorporate the provided context where relevant to make the questions more specific and meaningful.


Do not add anything else in the output.
\end{lstlisting}
\onecolumn
\clearpage
\section{Preference Evaluation Results}
\label{app:preference-category-eval-results}

\begin{figure*}[htp]
    \centering
    \begin{minipage}{1\textwidth}
        \centering
        \includegraphics[width=1\linewidth]{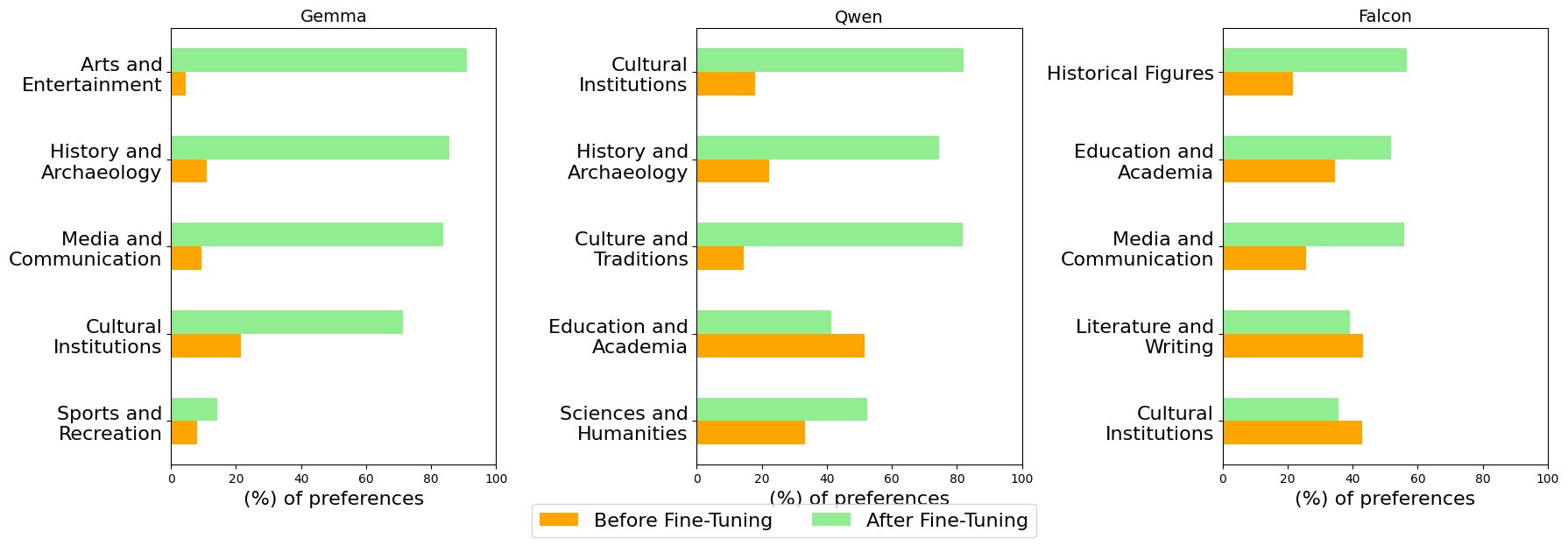}
        \caption*{(a) \texttt{CultSet}}
    \end{minipage}
    
    \begin{minipage}{1\textwidth}
        \centering
        \includegraphics[width=1\linewidth]{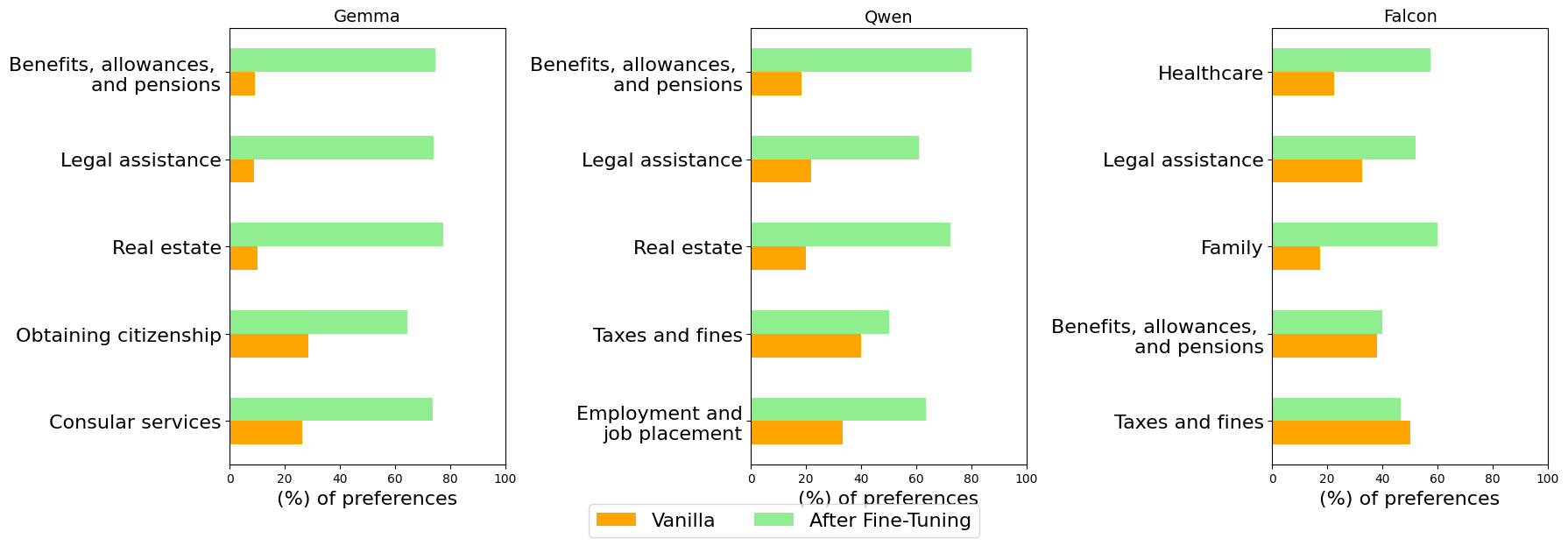}
        \caption*{(b) \texttt{GovSet}}
    \end{minipage}

    \caption{Category-wise comparison of preference evaluation across datasets: (a) \texttt{CultSet} and (b) \texttt{GovSet}.}
    \label{fig:category-wise-eval}
\end{figure*}

\clearpage
\onecolumn

\section{MURI Dataset Samples}
\label{appx:muri}
\begin{table*}[ht!]
    
    \centering
    \resizebox{0.8\linewidth}{!}{
    \begin{tabular}{|p{5cm}|p{5cm}|p{5cm}|}
        \hline
        \textbf{Instruction} & \textbf{Output} & \textbf{Issue} \\
        \hline
        \foreignlanguage{russian}{Morangis Франциядағы коммуна туралы мәлімет беру} & \foreignlanguage{russian}{Моренгем () — Франциядағы коммуна. Нор-Па-де-Кале аймағына қарасты Па-де-Кале департаментінде орналасқан. Сент-Омер округінің құрамына енеді. Алып жатқан жер аумағы 9,98 км² шамасында. Коммунаның INSEE коды — 62592, пошта индексі — 62910. Демографиясы 2009 жылғы мәліметтер бойынша тұрғындарының саны 485 адамды құрады. Халық тығыздығы — 48,6 адам/км². Тұрғындарының жас шамасы және жынысы бойынша құрамы (2006)...} &\foreignlanguage{russian}{Inconsistent use of Kazakh trans-literation, wrong verb's tense } \\
        \hline
        \foreignlanguage{russian}{Бірнеше танымал онлайн құмар ойындар веб-сайттарын көрсетіңіз.} &
        \foreignlanguage{russian}{***Құмар флеш покер ойындары – Қасқырлар мен қой ақша шығару ойын. | Онлайн покер 25. | Карьер ойындары солитей шаяны | Рулетка казино ойындарын ойнаңыз November 24, 2019 Қалай онлайн казино алдау Ханшайым академиясында ойнау 0 Comments 196 Казино флеш ойындар онлайн Ойын Real steel Android көп ақша жүктеп. VIP карталар казино Құмар ойындар онлайн сайт web. Казино бақылау Флорида. Казино flash player <...> Ақша сатып алу Android үшін ойындар. Sandia casino. Покер ақша тіркеу ойын. Покер-рум avi casino. Амазинг рп казино ойын тактикасы. Флорида-казино покер. Antiguo casino ponce Пуэрто-Рико. ...} &
        Inconsistent usage of Kazakh transliteration, markdown in the raw text\\
        \hline
        \foreignlanguage{russian}{Бердібек Сапарбаев Ақтөбе облысында «Жастар жылы» туралы не деді?} & 
        \foreignlanguage{russian}{Бердібек Сапарбаев Ақтөбе облысы Жастар жылын қалай өткізетінін айтты | BAQ.kz 28 Қаңтар 2019, 08:35 1857 | 1857 Астана, BAQ.KZ. Ақтөбе облысының әкімі Бердібек Сапарбаев 25 қаңтар, жұма күні Мемлекет басшысының Жастар жылы аясындағы тапсырмаларының орындалу мәселелерін талқылау үшін облыстағы жастар қозғалысының белсенділерімен кездесті.  ...} &
        Markdown and meta-data in the raw text \\
        \hline
    \end{tabular}
    }
    \caption{Samples from MURI dataset.}
    \label{tab:muri_samples}
\end{table*}
\clearpage

\section{Distribution of instruction types}

\begin{figure*}[ht]
    \centering
    \begin{minipage}[b]{0.45\textwidth}
        \centering
        \includegraphics[scale=0.34]{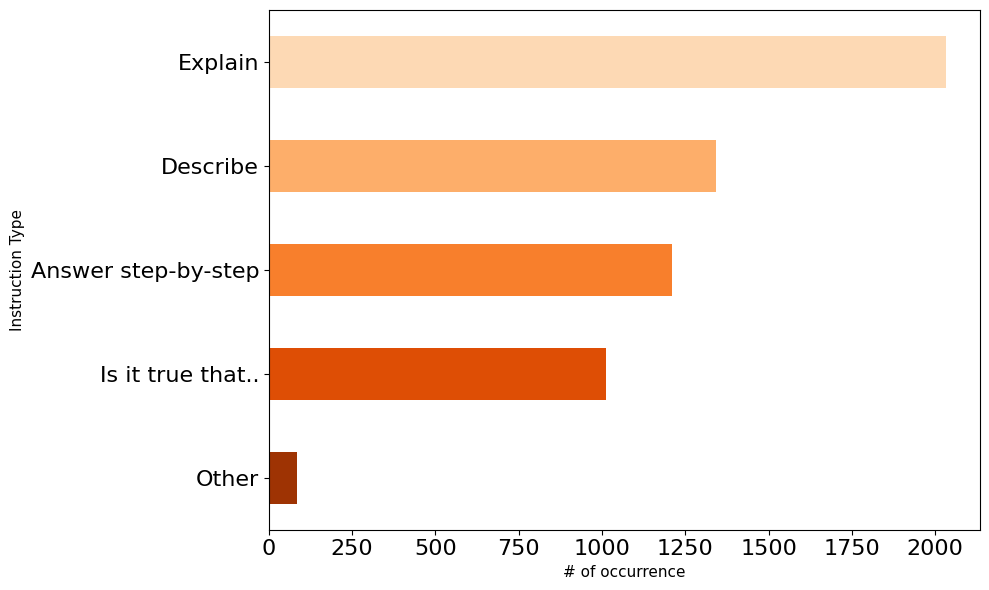}
        \caption*{  (a) \texttt{GovSet}  train instruction types}
        \includegraphics[scale=0.34]{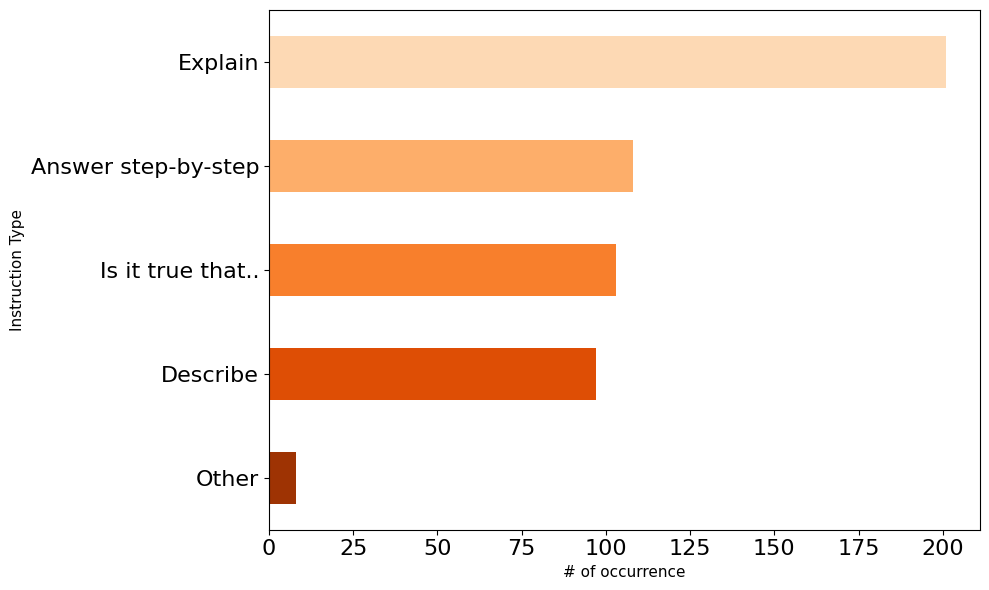}
        \caption*{  (b) \texttt{GovSet} test  instruction types}
    \end{minipage}
    \hfill
    \begin{minipage}[b]{0.45\textwidth}
        \centering
       \includegraphics[scale=0.34]{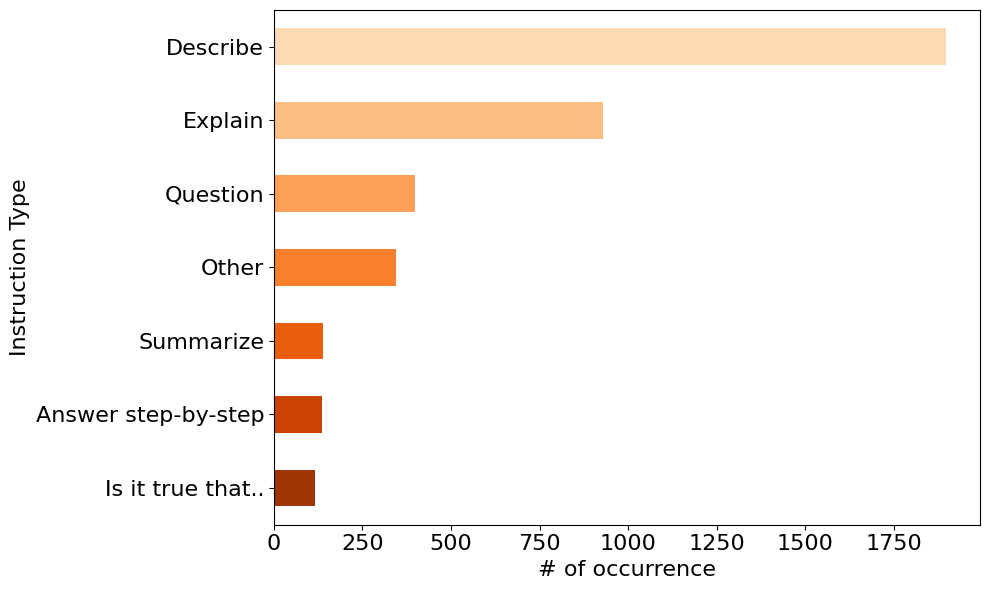}
        \caption*{  (c) \texttt{CultSet} train instruction types}
        \includegraphics[scale=0.34]{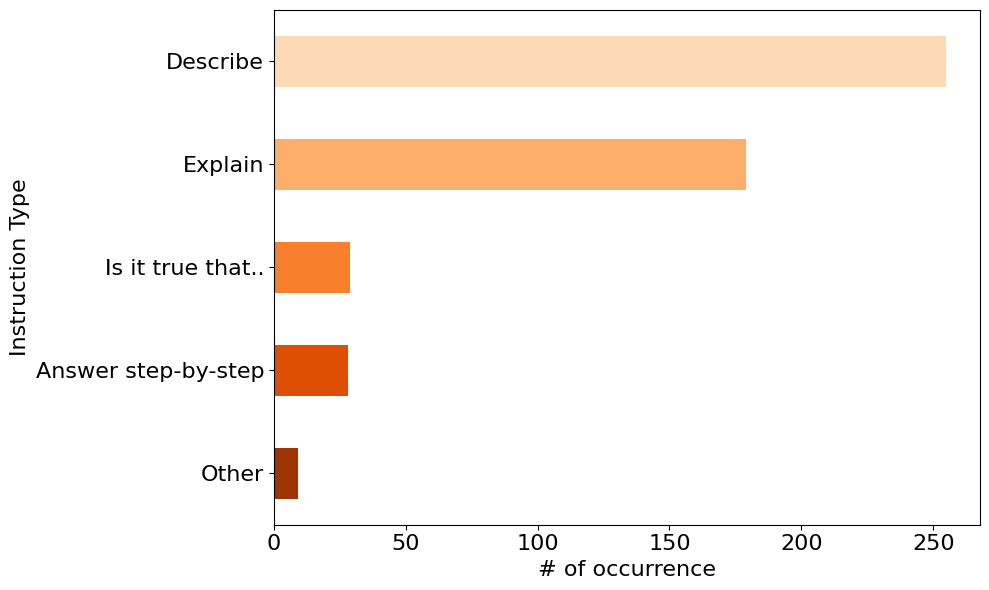}
        \caption*{  (d) \texttt{CultSet} test instruction types}
    \end{minipage}
    \caption{Instruction types distribution.}
    \label{fig:task_types}
\end{figure*}

\clearpage
\section{Preliminary Study}
\label{sec:prelimanary}
\FloatBarrier

\subsection{Human Evaluation for Preliminary Study Rubric}
\FloatBarrier

\begin{table*}[h]
\centering
\renewcommand{\arraystretch}{1.5} 
\scriptsize 
\begin{tabular}{>{\centering\arraybackslash}p{1.5cm}|>{\centering\arraybackslash}p{3.5cm}|>{\centering\arraybackslash}p{3.5cm}|>{\centering\arraybackslash}p{3.5cm}}
\hline
\textbf{Score} & \textbf{Correctness} & \textbf{Fluency} & \textbf{Completeness} \\ \hline
1 & Highly inaccurate, incorrect, or misleading information. & Very poor fluency, riddled with errors, making it difficult to read. & Very incomplete, with critical information missing, making it unusable. \\ \hline
2 & Significant factual or logical errors that impact the correctness of the instruction/question. & Multiple errors that hinder readability or cause confusion. & Significant omissions that make the instruction/question incomplete or difficult to interpret.  \\ \hline
3 & Noticeable errors in facts or logical flow, but the general meaning remains clear.
& Noticeable errors in grammar, spelling, or structure, but the text is still understandable. & Noticeable gaps in information or coverage that leave the instruction/question lacking. \\ \hline
4 & Minor factual inaccuracies or inconsistencies that do not affect overall understanding. & Minor grammatical or stylistic errors that do not significantly affect readability. & Slightly incomplete, with minor missing details that do not impact overall understanding. \\ \hline
5 & Fully correct and consistent with Kazakh cultural and governmental contexts, no factual or logical errors. & Perfect fluency, no errors in grammar, spelling, or sentence structure. The text reads smoothly and naturally. & Fully complete, no information is missing, and the instruction/question thoroughly covers the context.\\ \hline

\end{tabular}
\caption{Human annotation rubric for preliminary studies.}
\label{tab:multilingual-issues}
\end{table*}

\FloatBarrier

\clearpage
\section{Annotation Guideline}
\label{app:annot-guide-appendix}
To ensure a high-quality and standardized format for instruction-output annotations, we provide clear guidelines throughout the annotation process. The annotators refine and edit the automatically generated instruction fine-tuning (IFT) pairs using Google Spreadsheets, with each annotator assigned an individual worksheet. A detailed explanation of each field is provided below.
\\
\begin{itemize}
    \item \textbf{ID:} A unique identifier assigned to each data entry.

\item \textbf{Category}: This column contain the category of each data 
\item \textbf{Instruction:} The automatically generated instruction from the original text.
    \item \textbf{Output:} The corresponding generated output.
    \item \textbf{Updated Instruction:} A revised version of the instruction that has been edited.
    \item \textbf{Updated Output:} The modified output.
    \item \textbf{Comment:} Additional notes from the annotator, including explanations of any modifications, uncertainties about correctness, or suggestions for further improvements.

\end{itemize}

\textbf{General Rules}
\begin{itemize}
    \item Annotators must verify the correctness of the generated IFT pairs by comparing them against the original text. If an instruction-answer pair appears ambiguous, incorrect, or not supported by the original text, it should be highlighted for further review. The author (either the project owner or designated reviewers) will assess and remove it if necessary.
    \item Instructions must be complete and contextually accurate. If an instruction refers for example to a place, governmental process, or any specific entity but does not explicitly name it, annotators must incorporate the missing details from the original text.
    \item Questions must be fluent in Kazakh and maintain formal language for governmental data. No literary expressions, dialectal variations, or informal language should be introduced into governmental instructions—formality must be preserved.
    \item When processing biographical information, ensure clarity in numerical dates. 
    \item If an instruction-answer pair is completely unrelated to the original text, annotators must review the original text, verify the discrepancy, and highlight it for correction.
    \item Annotators are required to provide progress updates every two days, ensuring that issues are addressed promptly.
\end{itemize}

\clearpage
\section{Annotation example}
\label{sec:annotation example}
\subsection{GovSet}
\begin{table*}[ht!]
\centering
\resizebox{\linewidth}{!}{
\scriptsize
\begin{tabular}{@{}p{0.08\textwidth}p{0.22\textwidth}p{0.22\textwidth}p{0.22\textwidth}p{0.22\textwidth}@{}}

\toprule
\multicolumn{1}{c}{\textbf{Type}} & 
\multicolumn{1}{c}{\textbf{Instruction}} & 
\multicolumn{1}{c}{\textbf{Output}} & 
\multicolumn{1}{c}{\textbf{Updated Instruction}} & 
\multicolumn{1}{c}{\textbf{Updated Output}} \\
\midrule
\\
\multicolumn{1}{c}{\textbf{G}} & 
\foreignlanguage{russian}{Неке қиғаннан кейін құжаттарды ауыстыру үдерісін түсіндіріңіз.}
 & \foreignlanguage{russian}{Некеге тұрғаннан кейін жұбайы, егер ол жұбайының/зайыбының тегін алса, неке қию туралы куәлікті алған сәттен бастап 30 күн ішінде жеке куәлікті ауыстыру қажет.} & \foreignlanguage{russian}{\textcolor{darkgreen}{Қазақстанда} неке қиғаннан кейінгі құжаттарды ауыстыру үдерісін түсіндіріңіз.} & \foreignlanguage{russian}{\textcolor{darkgreen}{Ең алдымен, адам некеден кейін} жұбайының тегін \textcolor{darkgreen}{қабылдаса}, жеке куәлігін 30 күн ішінде ауыстыруы қажет.} \newline \\
 \\
 & Explain the process of changing documents after marriage. & After marriage, if the spouse takes the husband's/wife's last name, the national ID must be changed within 30 days from the date the marriage certificate is received. & Explain the process of changing documents after marriage in Kazakhstan. & First of all, if a person adopts their spouse's surname after marriage, they must change their ID within 30 days.\\
 \\
\hline
\\
\multicolumn{1}{c}{\textbf{G}} &
\foreignlanguage{russian}{Қазақстанда халықаралық рейсте жоғалған багаж үшін өтемақы төленетіні рас па?} & \foreignlanguage{russian}{Иә, халықаралық рейсте жоғалған багаж үшін өтемақы Қазақстан қол қойған екі Конвенцияның – Варшава мен Монреальдың ережелеріне, сондай-ақ компаниялардың ішкі ережелеріне сәйкес төленеді.} & \foreignlanguage{russian}{Қазақстанда халықаралық рейсте жоғалған багаж үшін өтемақы төлейтіні қаншалықты рас?} & \foreignlanguage{russian}{\textcolor{darkgreen}{Қазақстанда} халықаралық рейсте жоғалған багаж үшін өтемақы \textcolor{darkgreen}{төлейтіні толықтай рас. Өтемақы} Қазақстан қол қойған екі Конвенцияның \textcolor{darkgreen}{(}Варшава мен Монреальдың ережелеріне\textcolor{darkgreen}{)} \textcolor{darkgreen}{және} компаниялардың ішкі ережелері \textcolor{darkgreen}{негізінде есептелініп} төленеді.} \newline \\
 & Is it true that compensation is paid in Kazakhstan for lost baggage on international flights? & Yes, compensation for lost baggage on international flights is paid according to the rules of the two Conventions signed by Kazakhstan – the Warsaw and Montreal Conventions – as well as in accordance with the internal rules of the companies. & Is it true that compensation is paid in Kazakhstan for lost baggage on international flights? & It is completely true that compensation is paid in Kazakhstan for lost baggage on international flights. The compensation is calculated and paid based on the two Conventions signed by Kazakhstan (the rules of the Warsaw and Montreal Conventions) and the internal regulations of the companies.\\
\hline
\\
\multicolumn{1}{c}{\textbf{G}} &
\foreignlanguage{russian}{Қазақстанда қандастарға азаматтық алудың шарттары туралы түсіндіріңіз.} & \foreignlanguage{russian}{ Қандастар Қазақстан аумағында тұру мерзіміне қарамастан жеңілдетілген тәртіпте азаматтықты ала алады. Олар бұрынғы азаматтықтан бас тарту туралы нотариат куәландырған өтініш, бекітілген нысандағы сауалнама-өтініш, шарттарды сақтау туралы міндеттеме, өмірбаян, жеке басын куәландыратын және адамның басқа елдің азаматтығына жататындығын растайтын құжаттың көшірмесі, қандастың куәлігі және төрт фотосурет секілді құжаттарды тапсыруы керек.} & \foreignlanguage{russian}{\textcolor{darkgreen}{Қандастар Қазақстан Республикасының азаматтығын алу үшін не істеу керек?}} & \foreignlanguage{russian}{\textcolor{darkgreen}{Қандастардың Қазақстан Республикасында тұрған} мерзіміне қарамастан \textcolor{darkgreen}{олардың азаматтық алуы жеңілдетілген тәртіппен жүзеге асады. Қандастар Қазақстан Республикасының азаматтығын алу үшін} бұрынғы азаматтықтан бас тарту туралы нотариат куәландырған өтініш, бекітілген нысандағы сауалнама-өтініш, шарттарды сақтау туралы міндеттеме, өмірбаян, жеке басын куәландыратын және адамның басқа елдің азаматтығына жататындығын растайтын құжаттың көшірмесі, қандастың куәлігі және төрт фотосурет секілді құжаттарды тапсыруы керек.}\\ \\
 & Explain the conditions for obtaining citizenship for oralmans (ethnic Kazakhs) in Kazakhstan. & Oralmans (ethnic Kazakhs) can obtain citizenship in a simplified manner regardless of the duration of their stay in Kazakhstan. They need to submit a notarized application to renounce their previous citizenship, a completed application form in the approved format, a commitment to comply with the conditions, an autobiography, a copy of a document confirming their identity and lack of foreign citizenship, a certificate of oralman status, and four passport-sized photos. & What should oralmans (ethnic Kazakhs) do to obtain citizenship of the Republic of Kazakhstan? & Oralmans (ethnic Kazakhs) can obtain citizenship in Kazakhstan regardless of the duration of their stay in a simplified manner. To obtain Kazakhstani citizenship, they must submit a notarized application to renounce their previous citizenship, a completed application form in the approved format, a commitment to comply with the conditions, an autobiography, a copy of a document confirming their identity and lack of foreign citizenship, a certificate of oralman status, and four passport-sized photos.\\
\hline
\end{tabular}}
\caption{Example of annotation for \texttt{GovSet}.}
\label{tab:localizing-kz-ru-ex}
\end{table*}
\clearpage

\subsection{CultSet}
\begin{table*}[ht!]
\centering
\resizebox{\linewidth}{!}{
\scriptsize
\begin{tabular}{@{}p{0.08\textwidth}p{0.22\textwidth}p{0.22\textwidth}p{0.22\textwidth}p{0.22\textwidth}@{}}

\toprule
\multicolumn{1}{c}{\textbf{Type}} & 
\multicolumn{1}{c}{\textbf{Instruction}} & 
\multicolumn{1}{c}{\textbf{Output}} & 
\multicolumn{1}{c}{\textbf{Updated Instruction}} & 
\multicolumn{1}{c}{\textbf{Updated Output}} \\
\midrule
\\
\multicolumn{1}{c}{\textbf{C}} & 
\foreignlanguage{russian}{Фильмнің қандай жетістіктері болды?}
 & \foreignlanguage{russian}{«Транссібір экспресі» фильмі 1978 жылы Ереванда өткен Бүкілодақтық кинофестивальда 'ең үздік шытырман оқиғалы фильм үшін' сыйлығын алды. Сондай-ақ, Карловы Варыда өткен 21-Халықаралық кинофестивальда Азия, Африка және Латын Америкасы елдерінің симпозиумы қазылар алқасының сыйлығына ие болды. 1978 жылы режиссер Оразбаев пен суретші Қарсақбаевқа Қазақстан Мемлекеттік сыйлығы берілді.} & \foreignlanguage{russian}{\textcolor{darkgreen}{"Транссібір экспресі" атты қазақстандық} фильмнің қандай жетістіктері болды?} & \foreignlanguage{russian}{«Транссібір экспресі» фильмі \textcolor{darkgreen}{1977 жылы Қазақфильм студиясы тарабынан шықты.} 1978 жылы ол Ереванда өткен Бүкілодақтық кинофестивальда \textcolor{darkgreen}{"Ең үздік авантюра фильмі үшін" сыйлығын алды. Бұдан гөрі}, Карловы Варыда өткен 21-Халықаралық кинофестивальда Азия, Африка және Латын Америкасы елдерінің симпозиумы қазылар алқасының сыйлығына \textcolor{darkgreen}{да} ие болды. 1978 жылы фильм режиссері Оразбаев пен суретшісі Қарсақбаевқа Қазақстан Мемлекеттік сыйлығы берілді.}\newline
 \\
 \\
& What achievements did the film have? & The film "Trans-Siberian Express" received the award for "Best Adventure Film" at the World Film Festival held in Yerevan in 1978. Additionally, it won a prize at the 21st International Film Festival in Karlovy Vary, as well as at the symposium of Asia, Africa, and Latin American countries. In 1978, the film's director, Orazbayev, and cinematographer Karasbayev were awarded the State Prize of Kazakhstan. & What achievements did the Kazakhstani film "Trans-Siberian Express" have? & The film "Trans-Siberian Express" was released in 1977 by the Kazakhfilm studio. In 1978, it received the award for "Best Adventure Film" at the World Film Festival in Yerevan. Later, it also won a prize at the 21st International Film Festival in Karlovy Vary and at the symposium of Asia, Africa, and Latin American countries. In 1978, the film's director, Orazbayev, and cinematographer Karasbayev were awarded the State Prize of Kazakhstan. \\
\\
\hline
\\
\multicolumn{1}{c}{\textbf{C}} &
\foreignlanguage{russian}{Қазақ халқының музыкалық мұрасы қалай дамыды?} & \foreignlanguage{russian}{Қазақ халқының музыкалық мұрасы ғасырлар бойы ауыздан-аузға, ұстаздан-шәкіртке жеткізіліп, көшпенді өмір салтының ықпалымен дамыды. Музыкалық дәстүрлер батырлық және лирикалық жанрдағы эпостарды қамтыды. Қазақ композиторлары әлемдік деңгейде танымал болып, ұлттық музыкалық мәдениетті дамытуда үлкен үлес қосты.} & \foreignlanguage{russian}{Қазақ халқының музыкалық мұрасы қалай дамыды?} & \foreignlanguage{russian}{Қазақ халқының музыкалық мұрасы ғасырлар бойы ауыздан-аузға, ұстаздан-шәкіртке жеткіз\textcolor{darkgreen}{ілген. Негізінен} көшпенді өмір салтының ықпалымен дамыды. \textcolor{darkgreen}{Қазақтың} музыкалық дәстүрі батырлық және лирикалық жанрдағы эпостарды қамтыды. \textcolor{darkgreen}{Қазіргі заманда} қазақ композиторлары әлемдік деңгейде танымал болып, ұлттық музыкалық мәдениетті \textcolor{darkgreen}{одан әрі дамытуға үлкен үлес қосып жүр.}} \newline \\
& How did the musical heritage of the Kazakh people develop? & The musical heritage of the Kazakh people developed over centuries through oral transmission, from master to apprentice, and was influenced by the nomadic lifestyle. Musical traditions included epic genres of heroic and lyrical works. Kazakh composers became renowned internationally and contributed significantly to the development of national musical culture. & How did the musical heritage of the Kazakh people develop? & The musical heritage of the Kazakh people developed over centuries through oral transmission, from master to apprentice, primarily influenced by the nomadic lifestyle. Kazakh musical traditions encompassed genres of epic heroic and lyrical works. In modern times, Kazakh composers have gained international recognition and continue to contribute greatly to the development and advancement of national musical culture. \\
\\
\hline
\\
\multicolumn{1}{c}{\textbf{C}} &
 \foreignlanguage{russian}{Киіз үйде түндік дегеніміз не екенін түсіндіріңіз.} & \foreignlanguage{russian}{Түндік — киіз үйдің шаңырағына жабылатын киіз, ол әрі жарық береді, әрі түтін шығуға мүмкіндік береді. Төртбұрышты түрде пішіліп, шаңырақ диаметірінен 50-60 сантиметрдей артық жасалады.} & \foreignlanguage{russian}{\textcolor{darkgreen}{Қазақ халқының} киіз \textcolor{darkgreen}{үйіндегі түндіктің} не екенін түсіндіріңіз.} & \foreignlanguage{russian}{Түндік — \textcolor{darkgreen}{қазақтың баспанасы} киіз үйдің шаңырағына жабылатын киіз. \textcolor{darkgreen}{Ол әрі ішке жарық өткізеді, әрі түтін шығуына мүмкіндік береді}. Төртбұрышты түрде пішіліп, шаңырақ диаметірінен 50-60 сантиметрдей артық \textcolor{darkgreen}{етіп} жасалады.} 
 \\
& Explain what "tundik" means in a Kazakh yurt. & The "tundik" is the felt cover placed on the top of the yurt, which provides insulation and allows smoke to escape. It is made slightly larger than the circular roof opening, with a diameter exceeding 50-60 centimeters. It is triangular in shape. & Explain what "tundik" means in the Kazakh yurt. & The "tundik" is the felt cover placed on the top of the Kazakh yurt. It serves as insulation and allows smoke to escape. It is triangular in shape and is made slightly larger than the circular roof opening, with a diameter exceeding 50-60 centimeters.\\
\hline
 \end{tabular}}
\caption{Example of annotation for \texttt{CultSet}.}
\label{tab:localizing-kz-ru-ex}
\end{table*}

\clearpage
\section{Annotator Comments}
\label{app:annot-comments}
\subsection{GovSet}
\begin{table*}[ht!]
\scriptsize
\begin{tabular}{@{}c@{\hspace{10pt}}p{0.88\textwidth}@{}}
\toprule
\textbf{Error Type} & \textbf{Annotator Comments} \\ 

\midrule 
S & The generated data constantly uses an unnatural sentence structure that sounds like this: \foreignlanguage{russian}{\textit{Сіз сәбиіңіздің дүниеге келуін күтудесіз және онлайн режимінде емханаға тіркелу үдерісін түсіндіріңізші.}} Here the word \foreignlanguage{russian}{\textit{және}} sounds extra and can be either eliminated or replaced with \foreignlanguage{russian}{\textit{бұл жағдайда.}} This example of a sentence with extra \foreignlanguage{russian}{\textit{және}} is common. \\
\midrule
G & Inconsistent verb forms in instructions. Like \foreignlanguage{russian}{\textit{түсіндіру}} instead of \foreignlanguage{russian}{\textit{түсіндіріңіз.}} \\
\midrule
G & The generated data has two verb forms like \foreignlanguage{russian}{\textit{тану}} and \foreignlanguage{russian}{\textit{танылу}}. If the first one is the action of subject, the second is an action that happens TO a subject/object. 
 \\
\midrule
S & There is a frequent unnatural sentence structure that sounds like this: \foreignlanguage{russian}{\textit{Бұл рас па: некеге тұрған адамдарға куәлік электрондық нысанда беріледі?}} Either the whole statement needs to be reconstructed or \foreignlanguage{russian}{\textit{ме}} should be added at the end of the sentence.
 \\
\midrule
G & The model frequently fails to change words to accusative case (add \foreignlanguage{russian}{\textit{– н/- ын/-ін}}). For example,  in the phrase \foreignlanguage{russian}{\textit{тегі өзгертуге бола ма}}, the correct form would be \foreignlanguage{russian}{\textit{тегін өзгертуге бола ма}}.\\
\midrule
G & The translator often does not differentiate \foreignlanguage{russian}{\textit{кезінде}} and \foreignlanguage{russian}{\textit{кезіндегі}}. For example, in the sentence \foreignlanguage{russian}{\textit{Некені бұзу кезінде құжаттарды ресімдеу туралы түсіндіріңіз}}, the correct word would be \foreignlanguage{russian}{\textit{кезіндегі}}. The same happens with \foreignlanguage{russian}{\textit{Қазақстанда and Қазақстандағы}}.
\\
\midrule
L & The model writes the word \foreignlanguage{russian}{\textit{айырлуы}} several times. It must be either \foreignlanguage{russian}{\textit{айырылуы}} or \foreignlanguage{russian}{\textit{айырылғаны}} correctly. 
 \\
\midrule
S & The model frequently writes a phrase \foreignlanguage{russian}{\textit{жүзеге асырылады}}, which sounds unnatural and must be \foreignlanguage{russian}{\textit{жүзеге асады}} instead. 
\\
\midrule
G & The model frequently used phrases like \foreignlanguage{russian}{\textit{кірмейді екенін}}, when the correct form for any verb there would be \foreignlanguage{russian}{\textit{кірмейтінін}}. \\
\midrule
G & The model constantly uses this phrasing: \foreignlanguage{russian}{\textit{дегеніміз не екенін түсіндіріңіз.}} It sounds unnatural and must be replaced either with \foreignlanguage{russian}{\textit{не екенін түсіндіріңіз}} or just \foreignlanguage{russian}{\textit{дегеніміз не?}} \\ 
\midrule
G & The model cannot differentiate question words like \foreignlanguage{russian}{\textit{ма, ме}} etc. It can write \foreignlanguage{russian}{\textit{міндетті ма?}}, which is used in verbal speech only. The correct form would be \foreignlanguage{russian}{\textit{міндетті ме?}} \\ 
\midrule
G & The model can miss letters. For instance, \foreignlanguage{russian}{\textit{мақсатары}} instead of \foreignlanguage{russian}{\textit{мақсаттары}}; it is also not connected with the original text because the word is not even used there. \\ 
\midrule
S & There is a common sentence structure that sounds like this: \foreignlanguage{russian}{\textit{Тұрмыстық зорлық-зомбылық жағдайында қандай медициналық көмек алу жолдарын айтыңыз.}} Either the word \foreignlanguage{russian}{\textit{қандай}} must be removed and \foreignlanguage{russian}{\textit{-ғы}} added to \foreignlanguage{russian}{\textit{жағдайында}} (\foreignlanguage{russian}{\textit{жағдайындағы}}), or the statement must be a question instead, like \foreignlanguage{russian}{\textit{қандай медициналық көмек алуға болады?}} \\ 
\midrule
G & \foreignlanguage{russian}{\textit{Әр түрлі}} must be written as \foreignlanguage{russian}{\textit{әртүрлі}}. \foreignlanguage{russian}{\textit{Марапат}} must be used instead of \foreignlanguage{russian}{\textit{награда}}. \\ 
\midrule
S & There are many sentences like \foreignlanguage{russian}{\textit{Шындық па, үйде арнайы әлеуметтік қызметтерді көрсету тоқтатылады, егер қызмет алушының өзі олардан бас тартса?}} The structure is wrong, \foreignlanguage{russian}{\textit{шындық па}} must be placed at the end!! And \foreignlanguage{russian}{\textit{егер}} etc. must be placed at the start. Any verb before \foreignlanguage{russian}{\textit{шындық па}} must end with \foreignlanguage{russian}{\textit{-тыны/-тіні}}. \\ 
\midrule
G & The model often writes \foreignlanguage{russian}{\textit{деп түсіндіріңіз}} in instructions. In different occasions it must be written differently, like \foreignlanguage{russian}{\textit{бола ма, түсіндіріңіз}} or \foreignlanguage{russian}{\textit{болатынын түсіндіріңіз}}. \\ 
\midrule
G & Incorrect translation for \foreignlanguage{russian}{\textit{процесс}}. It should be translated as \foreignlanguage{russian}{\textit{үдеріс/үрдіс}}. \\ 
\midrule
L & Sentences like \foreignlanguage{russian}{\textit{Жеке куәлікті алғаш алу үшін қандай құжаттар қажет екені рас па?}} are common in “Is it true that” instructions. Either \foreignlanguage{russian}{\textit{қандай}} must be removed or replaced with a list of documents. \\ 
\midrule
L & The model generates a phrase \foreignlanguage{russian}{\textit{өтініш жасау}}; the more common version would be \foreignlanguage{russian}{\textit{өтініш беру}} or \foreignlanguage{russian}{\textit{өтініш жазу}}. \\
\midrule
G & The model sometimes has a hard time differentiating active and passive positions of subjects. For example, \foreignlanguage{russian}{\textit{тіркеу куәлігі немесе әскери билет жоғалтқанда}} puts the verb as if the subject is passive, whereas the correct form would be \foreignlanguage{russian}{\textit{жоғалғанда}} (because the subjects are in active form). This happens with different verbs throughout the document. \\
\midrule
G & Sometimes the answers that the model generates for instructions are not even connected to the question. For example, the answer for \foreignlanguage{russian}{\textit{Өтініш беруші ұсынылған ақпарат үшін қашан жауап береді деген айтылды ма?}} is \foreignlanguage{russian}{\textit{Иә, өтініш берушілер ұсынылған ақпарат үшін өздері жауап береді деп айтылды.}} The word \foreignlanguage{russian}{\textit{өздері}} as an answer is unrelated to the question \foreignlanguage{russian}{\textit{қашан}}. \\
\midrule
L & The model frequently generates phrases like \foreignlanguage{russian}{\textit{осы мәтін бойынша}}, \foreignlanguage{russian}{\textit{бұл деректерге қарағанда}}, which implies that there is a certain text to rely on. \\
\midrule
L & The model sometimes pronounces \foreignlanguage{russian}{\textit{Беларусь}} as \foreignlanguage{russian}{\textit{Беларуссия}}, which is not correct even unofficially. For formal texts, \foreignlanguage{russian}{\textit{Беларусь}} should be used, or at least \foreignlanguage{russian}{\textit{Белоруссия}}. \\
\midrule
L & The model frequently uses the Russian word \foreignlanguage{russian}{\textit{заем}}, even though there are Kazakh translations for it, such as \foreignlanguage{russian}{\textit{несие}} or \foreignlanguage{russian}{\textit{қарыз}}. This happens because of the original text. \\
\midrule
S & Sometimes the model generates sentence structures like \foreignlanguage{russian}{\textit{Бұл рас па: Азамат 16 жасқа толғанда Қазақстанның әрбір азаматы жеке куәлік алуға міндетті екенін?}} Firstly, the word \foreignlanguage{russian}{\textit{екенін}} in all these sentences is grammatically incorrect here and must be \foreignlanguage{russian}{\textit{екені}}. Additionally, since this sentence structure is very frequent, it must be reconstructed to place \foreignlanguage{russian}{\textit{бұл рас па}} at the end. For example, it should sound like this: \foreignlanguage{russian}{\textit{16 жасқа толғанда Қазақстанның әрбір азаматы жеке куәлік алуға міндетті екені рас па?}} \\
\midrule
S & There are also sentences that start with \foreignlanguage{russian}{\textit{шынайы ма екенін анықтаңыз}} or \foreignlanguage{russian}{\textit{шындық па анықтаңыз}}. These phrases must be rewritten as either \foreignlanguage{russian}{\textit{шындық па, соны анықтаңыз}} or just \foreignlanguage{russian}{\textit{шын ба}}/\foreignlanguage{russian}{\textit{шындық па}} without the \foreignlanguage{russian}{\textit{анықтаңыз}} part. \\
\midrule
G & The model cannot identify where it is better to not use \foreignlanguage{russian}{\textit{табыс септік}}. For example, \foreignlanguage{russian}{\textit{пәтерді іздеу}}/\foreignlanguage{russian}{\textit{пәтерді іздегенде}} is used frequently, but the common and natural way must be without the \foreignlanguage{russian}{\textit{-ді}} suffix. \\

\bottomrule
\end{tabular}
\caption{Selected annotator comments highlighting issues in \texttt{GovSet}. Error types are categorized as follows: L – Lexical errors, S – Structural errors, G – Grammatical errors.}
\label{table:annotator_comments_gov}
\end{table*}

\clearpage
\subsection{CultSet}
\begin{table*}[ht]
\scriptsize
\begin{tabular}{@{}c@{\hspace{10pt}}p{0.88\textwidth}@{}}
\toprule
\textbf{Error Type} & \textbf{Annotator Comments} \\ 
\midrule 
G & Inconsistent verb forms are present in the generated data. For example, \foreignlanguage{russian}{\textit{түсіндір}} is used instead of \foreignlanguage{russian}{\textit{түсіндіріңіз}}. \\
\midrule
G & The generated data includes phrases such as "\foreignlanguage{russian}{\textit{... туралы сипаттаңыз}}," which is grammatically incorrect because \foreignlanguage{russian}{\textit{туралы}} is redundant. \\
\midrule
S & Phrases like "\foreignlanguage{russian}{\textit{... қалай сипатталады?}}" appear awkward and imply the presence of external text for context. \\
\midrule
L & The word \foreignlanguage{russian}{\textit{карьера}} is used instead of the correct Kazakh term \foreignlanguage{russian}{\textit{мансап}}. \\
\midrule
G & The word \foreignlanguage{russian}{\textit{әртүрлі}} is written as \foreignlanguage{russian}{\textit{әр түрлі}}, which is inconsistent with standard formatting. \\
\midrule
G & There is confusion regarding whether \foreignlanguage{russian}{\textit{-мен}} (\foreignlanguage{russian}{\textit{-бен}}, \foreignlanguage{russian}{\textit{-пен}}) should be attached to or separated from nouns. \\
\midrule
G & The form of \foreignlanguage{russian}{\textit{жылдары}} is inconsistent, alternating between \foreignlanguage{russian}{\textit{жылы}}, \foreignlanguage{russian}{\textit{жылдары}}, or abbreviated forms like \foreignlanguage{russian}{\textit{ж.}} or \foreignlanguage{russian}{\textit{жж.}}. \\
\midrule
L & Russian words, such as \foreignlanguage{russian}{\textit{награда}}, \foreignlanguage{russian}{\textit{карьера}}, and \foreignlanguage{russian}{\textit{музей}}, are used even though Kazakh equivalents exist. \\
\midrule
L & The word \foreignlanguage{russian}{\textit{артис}} is used instead of the correct Kazakh term \foreignlanguage{russian}{\textit{әртіс}}. \\
\midrule
S & Words like \foreignlanguage{russian}{\textit{бастап}} and \foreignlanguage{russian}{\textit{бері}} are omitted when referring to years. For example, \foreignlanguage{russian}{\textit{1967 жылдан ол бұл атаққа ие болды}}. \\
\midrule
G & The phrase \foreignlanguage{russian}{\textit{туралы түсіндіріңіз}} is unnatural and often incorrect. It should be replaced with \foreignlanguage{russian}{\textit{туралы айтыңыз}} or \foreignlanguage{russian}{\textit{айтып беріңіз}}. \\
\midrule
G & The phrase \foreignlanguage{russian}{\textit{жеңіл атлетші}} is used correctly, but \foreignlanguage{russian}{\textit{ауыр атлет}} is written without the suffix \foreignlanguage{russian}{\textit{-ші}}. It should be standardized as \foreignlanguage{russian}{\textit{атлетші}}. \\
\midrule
L & Shortened names and surnames are written without spaces. For example, \foreignlanguage{russian}{\textit{Ы.Алтынсарин}} instead of \foreignlanguage{russian}{\textit{Ы. Алтынсарин}}. \\
\midrule
L & The phrase \foreignlanguage{russian}{\textit{рөл ойнады}} is a direct translation from Russian and should be replaced with \foreignlanguage{russian}{\textit{рөл орындады}} or \foreignlanguage{russian}{\textit{рөл сомдады}}. \\
\midrule
L & Years are written in abbreviated forms, such as \foreignlanguage{russian}{\textit{1973-78}}, omitting \foreignlanguage{russian}{\textit{19}} in the second part. \\
\midrule
L & Names of \foreignlanguage{russian}{\textit{ру}}, tribes (\foreignlanguage{russian}{\textit{тайпа}}), and \foreignlanguage{russian}{\textit{жүз}} are written in lowercase, which is inconsistent with standard capitalization rules. \\
\midrule
L & The word \foreignlanguage{russian}{\textit{рөл}} is occasionally misspelled as \foreignlanguage{russian}{\textit{рол}} or \foreignlanguage{russian}{\textit{роль}}. \\
\midrule
G & Verbs ending with \foreignlanguage{russian}{\textit{-ып}} or \foreignlanguage{russian}{\textit{-іп}} are used incorrectly, leading to unrelated sentences. For example, in the sentence \foreignlanguage{russian}{\textit{Бұл ойын үлкен тойларда арнайы жүлде тағайындалып өткізіліп, оның түрлері көптеген}}, the part after \foreignlanguage{russian}{\textit{өткізіліп}} is unrelated. The correct form would be \foreignlanguage{russian}{\textit{Бұл ойын үлкен тойларда арнайы жүлде тағайындалып өткізіледі; оның түрлері көп}} or \foreignlanguage{russian}{\textit{Бұл ойын үлкен тойларда арнайы жүлде тағайындалып өткізіліп, қазақ халқында әлі де жүргізіліп келеді}}. \\
\midrule
S & Words occasionally appear in the Latin alphabet even though they are not written in Latin in the original text. \\
\midrule
G & Sentences like \foreignlanguage{russian}{\textit{Безендірудің көркемдеу құралдарын қандай өнер салалары пайдаланады деп түсіндіріңіз}} are included. The common error is the use of \foreignlanguage{russian}{\textit{пайдаланады деп түсіндіріңіз}}, which should be replaced with \foreignlanguage{russian}{\textit{пайдаланатынын түсіндіріңіз}}. \\
\midrule
G & The phrase \foreignlanguage{russian}{\textit{жер аударылды}} is used for referring to a person. If it describes someone moving places, it should be corrected to \foreignlanguage{russian}{\textit{жер аударды}}. \\
\midrule
S & The structure of the questions deviates from the natural structure of the Kazakh language. For example: \foreignlanguage{russian}{\textit{Бұл рас па, АХҚО қатысушыларының қызметін реттеу Англия мен Уэльстің жалпы құқық қағидаттары негізінде жүргізіледі?}} \\
\midrule
S & The questions ask for information not mentioned in the text, meaning the answer to the question is not present in the text. \\
\midrule
S  & The structure of the questions is entirely incorrect. For example: \foreignlanguage{russian}{Астана Халықаралық Қаржы Орталығы қандай компаниялар үшін реттеуге икемді тәсіл қолданады?}
The correct version: \foreignlanguage{russian}{Астана Халықаралық Қаржы Орталығы қандай компанияларды реттеуге мүмкіндік береді?} 
\\
\midrule
L & The word \foreignlanguage{russian}{\textit{ол}} is overused in the generated data. \\
\bottomrule
\end{tabular}
\caption{Selected annotator comments highlighting issues in \texttt{CultSet}. Error types are categorized as follows: L – Lexical errors, S – Structural errors, G – Grammatical errors.}
\label{table:annotator_comments_cult}
\end{table*}

\section{Categories Explained}
\label{sec:category_distribution}

\begin{table*}
\scriptsize
    \centering
    \begin{tabular}{p{0.25\textwidth}p{0.7\textwidth}}
    \toprule
\textbf{Category} & \textbf{Description} \\  
\midrule
Legal Assistance & Services related to providing legal advice, resolving disputes, or ensuring compliance with laws in areas such as property rights, labor law, or civil matters. \\  \midrule
Healthcare & Includes services for accessing healthcare facilities, obtaining health insurance, and managing public health initiatives, including vaccinations and medical records. \\  \midrule
Real Estate & Covers registration of property rights, encumbrances, and the legal transfer of property ownership. \\  \midrule
Education & Involves services for school and university admissions, recognition of diplomas, and access to government scholarships or grants. \\  \midrule
Family & Services addressing marriage registration, child adoption, and other family-related matters like child welfare and maternity benefits. \\  \midrule
Taxes and Fines & Includes filing taxes, obtaining tax statements, and paying administrative fines or penalties. \\  \midrule
Business Registration & Covers the process of starting, registering, or closing businesses, including obtaining licenses and permits. \\  \midrule
Benefits, Allowances, and Pension & Government services related to social welfare benefits, pensions, and financial aid for citizens. \\  \midrule
Employment and Job Placement & Services that facilitate job searches, vocational training, and unemployment benefits. \\  \midrule
Driving and Transportation & Includes services for driver’s license issuance, vehicle registration, and public transportation systems. \\  \midrule
ID Card, Passport, and Registration & Involves obtaining or renewing identity documents, passports, and residency registration. \\  \midrule
Industry & Services supporting industrial development, including subsidies, licenses, and monitoring compliance with regulations. \\  \midrule
Financial Literacy & Government initiatives to improve citizens' understanding of personal finance, savings, and investment. \\  \midrule
Consular Services & Services provided to citizens abroad, such as visa issuance, document legalization, and emergency support. \\  \midrule
Travel, Living Abroad, and Emigration & Addresses issues like emigration permits, citizenship renunciation, and support for expats. \\  \midrule
Licensing and Accreditation & Involves issuing licenses for various professional or business activities and accrediting institutions. \\  \midrule
Obtaining Citizenship & Covers processes for naturalization, renouncing, or restoring Kazakh citizenship. \\  \midrule
Special Services & Includes unique or niche government-provided services that do not fall into other categories. \\  \midrule
Military Registration & Services related to military service registration, deferments, and exemptions. \\  \midrule
Population & Addresses population census activities, demographic statistics, and citizen registration systems. \\  \midrule
Religion & Involves registration and oversight of religious organizations and facilitation of religious freedom policies. \\  
\bottomrule
\end{tabular}
    \caption{Category explanation for \texttt{GovSet}.}
    \label{tab:category_exp_gov}
\end{table*}

\begin{table*}
\scriptsize
\begin{tabular}{p{0.25\textwidth} p{0.7\textwidth}}
    \toprule
    \textbf{Category} & \textbf{Description} \\ 
    \midrule
    Literature and Writing & This category includes Kazakh literature, poetry, and contemporary works. \\ \midrule
    Culture and Traditions & Covers Kazakh cultural practices, including traditional music, dances, festivals, and customs. \\ \midrule
    Media and Communication & Includes Kazakhstan’s media landscape, featuring national TV channels, news outlets. \\ \midrule
    Arts and Entertainment & Encompasses visual arts, theater, film, and music. \\ \midrule
    Education and Academia & Covers the education system in Kazakhstan, including primary to higher education. \\ \midrule
    History and Archaeology & Includes the exploration of Kazakhstan’s ancient sites and the history of the Kazakh people. \\ \midrule
    Politics & Encompasses the political landscape of Kazakhstan, including its government, political parties, and international relations, with a focus on its role in Central Asia. \\ \midrule
    Sciences and Humanities & Includes research and advancements in fields such as environmental science, economics, sociology, and the study of Kazakh identity and history. \\ \midrule
    Historical Figures & This category celebrates important figures in Kazakhstan’s history. \\ \midrule
    Sports and Recreation & Covers popular sports in Kazakhstan like boxing, wrestling, football, traditional Kazakh sports and famous Kazakhstani athletes. \\ \midrule
    Culinary and Food & This includes Kazakhstan's traditional cuisine, such as "beshbarmak" and "kymyz". \\ \midrule
    Health and Medicine & Covers public health systems, advancements in medical care and renowned Kazakh doctors. \\ \midrule
    Business and Industry & This category encompasses Kazakhstan’s growing economy, with a focus on industries like oil and gas, agriculture, and its expanding IT sector. \\ \midrule
    Cultural Institutions & Includes museums, galleries, theaters, and cultural centers in Kazakhstan, such as the National Museum of Kazakhstan and the Astana Opera. \\ \midrule
    Transportation and Infrastructure & Covers the country’s infrastructure development, such as its extensive rail system, airports. \\ \midrule
    Nature and Environment & Focuses on Kazakhstan’s diverse landscapes, such as the steppes, mountains, and lakes. \\ \midrule
    Miscellaneous & A catch-all category that could include any other cultural, social, or historical aspects of Kazakhstan not covered by other categories. \\ 
    \bottomrule
\end{tabular}
    \caption{Category explanation for \texttt{CultSet}.}
    \label{tab:category_exp_culture}
\end{table*}

\section{Example of created data}
\label{sec:exampleofcreatedata}
\begin{table*}[ht!]
\scriptsize
\centering
\scalebox{0.8}{
\begin{tabular}{p{0.15\textwidth}p{0.2\textwidth}p{0.7\textwidth}}
\toprule
\textbf{Category} & \textbf{Instruction} & \textbf{Output} \\ 
\midrule
Benefits, allowances, and pensions & \foreignlanguage{russian}{ Қазақстанда зейнетақы төлемдерін есептеудің барлық қадамдарын тізбектеңіз. } & \foreignlanguage{russian}{ 1. Табыс қатарынан 36 күнтізбелік ай ішіндегі жалпы соманы алыңыз.
2. Оны отыз алтыға бөліңіз.
3. Жасына байланысты зейнетақы төлемдерін есептеуге арналған кіріс, республикалық бюджет туралы заңда тиісті қаржы жылына белгіленген айлық есептік көрсеткіштің 55 еселенген мөлшерінен аспауға тиіс. } \\ 
\\
\\
 & List all the steps for calculating pension payments in Kazakhstan. & 1. Take the total amount of income for 36 calendar months in a row.
2. Divide it by thirty-six.
3. Income for calculating pension payments by age should not exceed 55 times the monthly calculation index established for the corresponding financial year by the Law on the Republican budget.
  \\
\midrule

Business registration & \foreignlanguage{russian}{ Бұл жоба қандай мақсатқа бағытталғанын түсіндіріңіз. } & \foreignlanguage{russian}{ "Zhas Project" - бұл әлеуметтік жобалар мен бизнес-идеяларды қаржыландыру арқылы осал топтарды, әсіресе ауылдық жерлерде тұратын жас қазақстандықтарды қолдауға бағытталған бірегей жоба. Жоба жастарды өз қоғамының әлеуметтік-экономикалық өміріне тартуға және нәтижелі жұмыспен қамтуға көмектеседі. } \\ 
\\
 & Explain the purpose of this project. & "Zhas Project" is a unique project aimed at supporting vulnerable groups, especially young Kazakhstanians living in rural areas, through the financing of social projects and business ideas. This project helps to involve young people in the socio-economic life of their society and ensures productive employment.  \\
\midrule

Consular services & \foreignlanguage{russian}{ Қазақстанның шетелдегі заңдастыру процесіне жеке және заңды тұлғаларға арналған құжаттардың ұқсастықтары мен айырмашылықтарын сипаттаңыз. } & \foreignlanguage{russian}{ Ұқсастықтары: Екі жағдайда да еркін нысандағы өтініш, заңдастыруға жататын құжат, және консулдық алымның төленгенін растайтын құжат талап етіледі. Айырмашылықтары: Жеке тұлға үшін жеке басын куәландыратын құжаттың көшірмесі қажет, ал заңды тұлға үшін уәкілетті тұлғаның куәландыратын құжаты қажет. Заңды тұлға үшін қосымша заңды тұлғаның мемлекеттік тіркелгенін (қайта тіркелгенін) растайтын құжат немесе заңды тұлғаның әрекет қабілеттілігін растайтын өзге де құжаттар талап етіледі. } \\ 
 & Describe the similarities and differences between documents for natural persons and legal entities in the legalization process of Kazakhstan abroad. & Similarities: in both cases, an application in any form, a document subject to legalization, and a document confirming the payment of the consular fee are required. Differences: for a natural person, a copy of an identity document is required, and for a legal entity, a certificate of an authorized person is required. For a legal entity, an additional document confirming the state registration (re-registration) of a legal entity or other documents confirming the legal capacity of a legal entity is required.  \\
\midrule

Driving and transportation & \foreignlanguage{russian}{ Транзиттік декларация көшірмелерінің талап етілетін жағдайын түсіндіріңіз. } & \foreignlanguage{russian}{ Транзиттік декларация ретінде пайдаланылатын құжаттар, мысалы, көліктік (тасымалдау) құжат, негізгі құжат болып табылады, ал басқа құжаттар оның ажыратылмас бөлігі ретінде қарастырылады. } \\ 
 & Explain the required status of the transit declaration copies. & Documents used as a transit declaration, for example, a transport (transportation) document, is the main document, while other documents are considered as its inalienable part.  \\
\midrule

Education & \foreignlanguage{russian}{ «Болашақ» бағдарламасы бойынша оқуға түсерде қандай құжаттарды алып келу керек? } & \foreignlanguage{russian}{ 1. Стипендиаттың жеке басын куәландыратын құжат. 2. Жұмыс беруші атынан оқу/тағылымдамадан өту шартына қол қоятын адамның өкілеттігін куәландыратын құжаттар. 3. Стипендиаттың ауылдық елді мекендерде тұрақты тұрғылықты жері туралы мәліметтер. } \\ 
\\
 & What documents should be brought when entering the Bolashak program? & 1. A document verifying the identity of the scholarship holder. 2. Documents certifying the authority of the person signing the studying/internship agreement on behalf of the employer. 3. Infromation on the permanent place of residence of the scholarship holder in rural settlements. \\
\midrule

Employment and job placement & \foreignlanguage{russian}{ «Бастау Бизнес» онлайн курсының ұзақтығы қанша күн болатынын сипаттаңыз. } & \foreignlanguage{russian}{ Оқу 14 күнтізбелік күннен аспайды. } \\ 
\\
 & Describe how many days the online course "Bastau Business" will last. & Training lasts no more than 14 calendar days.  \\
\midrule
Family & \foreignlanguage{russian}{ Алимент төлеу рәсімі туралы айтып беріңіз. } & \foreignlanguage{russian}{ Алимент төлеу рәсімі – бұл ата-аналардың баланы асырап-бағуға арналған төлемдер бойынша міндеттемелерін орындауға байланысты процесс. Бұл ең көп таралған алимент жағдайы ретінде белгілі. } \\ 
\\
\\
 & Tell about the procedure for paying alimony. & The alimony payment procedure is a process related to the fulfillment by parents of their child-support obligations. This is known as the most common case of alimony.  \\
\midrule

Financial literacy & \foreignlanguage{russian}{ АХҚО-ның жасыл қаржыны дамыту бойынша жүргізген жұмыстарына нелер жатады? } & \foreignlanguage{russian}{ АХҚО жасыл қаржыны дамыту үшін әлеуетті эмитенттерге, инвесторларға және нарық ойыншыларына алғашқы көмек көрсетеді. } \\ 

 & What is the work of the AIFC on the development of green finance? & The AIFC provides first aid to potential issuers, investors and market players for the development of green finance.  \\
\midrule
Healthcare & \foreignlanguage{russian}{ Емханаға қалай жазылуға болады? } & \foreignlanguage{russian}{ Жеке онлайн-сервисті пайдаланып немесе емхананың тіркеу бөліміне жүгініп, дәрігердің қабылдауына жазылуға болады. } \\

 & How can you make an appointment at the hospital? & You can make an appointment with a doctor using a personal online service or by contacting the registry office of the hospital.  \\ 
\midrule

ID card, passport, and registration & \foreignlanguage{russian}{ Жол жүру құжатын алу үшін қандай құжаттардың біреуін көрсету қажет? } & \foreignlanguage{russian}{ Жол жүру құжатын алу үшін қажетті құжаттардың біреуін көрсету қажет: туу туралы акт жазбасы, туу туралы куәлік, азаматтығы жоқ адамның куәлігі, шетелдік паспорт, босқын куәлігі. } \\ 

 & What documents do you need to show to obrain a travel document? & To obtain a travel document, it is necessary to show one of the necessary documents: a birth certificate record, a birth certificate, a certificate of absence of citizenship, a foreign passport, a refugee certificate.  \\
\bottomrule
\end{tabular} }
\caption{Example of created instructions on \texttt{GovSet}.}
\label{tab:example_govset}
\end{table*}

\begin{table*}[b!]
\scriptsize
\centering
\scalebox{0.8}{
\begin{tabular}{p{0.15\textwidth}p{0.2\textwidth}p{0.7\textwidth}}
\toprule
Category & Instruction & Output \\ 
\midrule

Industry & \foreignlanguage{russian}{Жеңіл өнеркәсіптің құрамында қандай қосалқы салалар бар?} & \foreignlanguage{russian}{Жеңіл өнеркәсіп 20-дан астам қосалқы салаларды қамтиды, олар тоқыма, тігін және былғары, тері, аяқ киім өндірістері.} \\ 

 & What sub-industries are part of the light industry? & Light industry includes more than 20 sub-industries, these are textile, sewing and leather, shoe industries.  \\
\midrule

Legal assistance & \foreignlanguage{russian}{Балалардың интернетте көп уақыт өткізуі рас па?} & \foreignlanguage{russian}{Иә, балалар мен жасөспірімдер интернетте көп уақыт өткізеді.} \\ 

 & Is it true that children spend a lot of time on the internet? & Yes, children and teenagers spend a lot of time on the internet.  \\
\midrule
Licensing and accreditation & \foreignlanguage{russian}{ Әуесқойлық аң аулауға куәлікті алу үшін қандай құжаттар қажет? } & \foreignlanguage{russian}{ Келесі құжаттар ұсынылуы тиіс: жеке куәлік, өлшемі 3х4см фотосурет (3 дана), аңшылық минимум курсын аяқтау сертификаты, 2 АЕК мөлшерінде мемлекеттік бажды төлеу туралы түбіртектің көшірмесі. } \\ 

 & What documents are needed to obtain a certificate for amateur hunting? & The following documents must be submitted: an identity card, a 3x4cm photo (3 pieces), a certificate of completion of the hunting minimum course, a copy of the receipt for payment of the state duty in the amount of 2 MCI.  \\
\midrule

Military registration & \foreignlanguage{russian}{ 2023 жылғы 16 қарашадан бастап әскери есепке алу қағидаларына қандай өзгерістер енгізілді? } & \foreignlanguage{russian}{ 2023 жылғы 16 қарашадан бастап бейбіт уақытта тіркеу туралы куәліктер беру тоқтатылды, ал әскери билеттер әскери қызметтің не құқық қорғау, арнаулы мемлекеттік органдарда қызметтің белгіленген мерзімінен өткен адамдарға ғана беріледі. } \\ 
\\
\\
 & What changes have been made to the rules of military registration since November 16, 2023? & Since November 16, 2023, the issuance of certificates of registration in peacetime has been suspended, and military tickets are issued only to persons who have passed the established period of military service or service in law enforcement, special state bodies.  \\
\midrule

Obtaining citizenship & \foreignlanguage{russian}{ Қазақстанның ЖОО-да оқитын этникалық қазақтар азаматтықты жеңілдетілген тәртіпте алады ма? } & \foreignlanguage{russian}{ Иә, Қазақстанның ЖОО-да оқитын этникалық қазақтар азаматтықты жеңілдетілген тәртіпте алады. } \\ \midrule
Population & \foreignlanguage{russian}{ 2023 жылы Қазақстанда қай этникалық топ халықтың көп бөлігін құрады? } & \foreignlanguage{russian}{ 2023 жылдың мәліметтер бойынша, қазақтар халықтың көп бөлігін құрайды (70,6\%). } \\ 

 & Do ethnic Kazakhs studying at universities in Kazakhstan receive citizenship in a simplified prodecure? & Yes, ethnic Kazakhs studying at universities in Kazakhstan receive citizenship in a simplified prodecure.  \\
\midrule

Real estate & \foreignlanguage{russian}{ Кәмелетке толмаған адамның өз мүлкін басқара алуы шындық па? } & \foreignlanguage{russian}{ Жоқ, кәмелетке толмаған пәтер иесі 18 жасқа толғанға дейін өз бетінше мүлкін басқара алмайды. } \\ 

 & Is it true that a minor can manage his/her property? & No, a minor apartment owner cannot independently manage his/her property until he/she reaches the age of 18.  \\
\midrule

Religion & \foreignlanguage{russian}{ Қазақстан халқының негізгі бөлігі қандай діндерді ұстанады? } & \foreignlanguage{russian}{ Елімізде халықтың негізгі бөлігі сунниттік исламды ұстанатын мұсылмандар және орыс православтық христиандықты ұстанушылар болып бөлінеді. } \\ 

 & What religions do the majority of the population of Kazakhstan practice? & In the country, the majority of the population is divided into Muslims professing Sunni Islam and adherents of Russian Orthodox Christianity.  \\
\midrule

Special services & \foreignlanguage{russian}{ Мемлекеттік органда қызметті алудағы қажетті құжаттарды түсіндіріңіз. } & \foreignlanguage{russian}{ Жеке басты куәландыратын, тұрақты тұрғылықты жері бойынша тіркелгенін растайтын құжаттар, мүгедектігі туралы және тиісті мемлекеттік ақпараттық жүйелерден мәліметтер қажет. Аталған құжаттар Мемлекеттік корпорацияның және жұмыспен қамту бөлімдерінің бөлімшелері арқылы ЭЦҚ-мен куәландырылған электрондық құжаттар нысанында алынады. } \\ 

 & List the necessary documents in receiving a service in a state body. & Documents confirming registration at the place of permanent residence, disability information and information from the relevant state data systems are required. These documents are obtained through the divisions of the State Corporation and Employment Departments in the form of electronic documents certified by e-signature.  \\
\midrule

Taxes and fines & \foreignlanguage{russian}{ Eqonaq жүйесінде шетелдік туристердің келу мерзімдерін сақтау үшін қандай құрал ұсынылады? } & \foreignlanguage{russian}{ Шетелдік туристердің келу мерзімдерін сақтау үшін Eqonaq жеке кабинетінің басты бетінде көші-қон анықтамалығы ұсынылған. } \\ 

 & What tool is offered in the Eqonaq system to comply with the deadlines for the arrival of foreign tourists? & To comply with the deadlines for the arrival of foreign tourists, a migration guide is provided on the main page of the Eqonaq personal account. \\
\midrule

Travel, living abroad, and emigration & \foreignlanguage{russian}{ Қазақстан Республикасының азамттығынан шығу үшін талап етілетін негізгі шарттарды түсіндіріңіз. } & \foreignlanguage{russian}{ Қазақстан Республикасының азамттығынан шығу үшін негізгі шарттар: азаматтың мемлекет алдында атқарылмаған міндеттері, қылмыстық жауапкершілікке тартылмауы, және азаматтықтан шығудың мемлекеттік қауіпсіздік мүдделеріне қайшы келмеуі тиіс. Егер осы шарттардың бірі бұзылса, азаматтықтан шығуға өтініш қабылданбайды. } \\ 

 & Explain the basic conditions required to renounce citizenship of the Republic of Kazakhstan. & The main conditions for renouncing the citizenship of the Republic of Kazakhstan are: unfulfilled duties of a citizen to the state, non-involvement in criminal liability, and non-renouncing citizenship should not contradict the interests of State Security. If one of these conditions is violated, the application for renunciation of citizenship will not be accepted.  \\
\bottomrule
\end{tabular} }
\caption{Example of created instructions on \texttt{GovSet}.}
\end{table*}

\begin{table*}[b!]
\scriptsize
\centering
\scalebox{0.8}{
\begin{tabular}{p{0.15\textwidth}p{0.2\textwidth}p{0.7\textwidth}}
\toprule
Category & Instruction & Output \\
\midrule
Arts and Entertainment & \foreignlanguage{russian}{Қазақ композиторы мен этномузыканты Еділ Сейілханұлы Құсайыновтың қандай атқарылымды еңбектері бар екенін түсіндіріңіз.} & \foreignlanguage{russian}{Еділ Сейілханұлы Құсайынов — «Қобыз» кафедрасының оқытушысы, түрлі ансамбльдер мен оркестрлерге арналған музыкалық шығармалардың, сондай-ақ «Көкбөрі-Күлтегін» балетінің, «Құрақ көрпе», «Келін» кинофильмдерінің музыкасының авторы. Ол «Алтай-Тарбағатай өңірінің сыбызғы күйлері» кітабының авторы, Қазақстан Республикасының Мәдениет Қайраткері атанған.} \\
 & Explain what are the works of the Kazakh composer and ethnomusician Edil Seilkhanuly Kusainov. & Edil Seilkhanuly Kusainov is an instructor at the "Kobyz" Department, the author of musical compositions for various ensembles and orchestras, as well as music for the "Kokbori-Kultegin" ballet, and the "Kurak Korpe", "Kelin" movies. He is the author of the book "Altai-Tarbagatai onirinin sybyzgy kuileri" (Sybyzgy Kuys of the Altai-Tarbagatai region), and a cultural figure of the Republic of Kazakhstan.  \\ 
\midrule
Business and Industry & \foreignlanguage{russian}{Қазақтың кен инженері мен бұрынғы Түпқараған ауданының әкімі Әлімғали Нұржанұлы Қапуовтың қызмет жолын сипаттаңыз.} & \foreignlanguage{russian}{Әлімғали Нұржанұлы Қапуов – кен инженері, бұрынғы Түпқараған ауданының әкімі. 1948 жылы Форт-Шевченко қаласында туған. ҚазПТИ-ді 1972 жылы бітірген. 1972-2005 жылдары әртүрлі қызметтер атқарған: Жетібай, «Ембімұнай» карьерінде, «Маңғыстау ұлутас» өндірістік бірлестігінде, «Маңғышлақ ауылқұрылыс-12» тресінде, «Маңғыстаумұнайқұрылыс» тресінде, «Өзен» фирмасында және тағы басқа орындарда қызметте болған. 2006 жылдан бастап «Құбыр» ЖШС-нің директоры. «Өзен мұнай кен орнына 40 жыл», «Қазақстан мұнайына 100 жыл» белгілерінің иегері.} \\
& Describe the career path of the Kazakh mining engineer and former Akim of Tupkaragan district Alimgali Nurzhanuly Kapuov. & Alimgali Nurzhanuly Kapuov - mining engineer, former Akim of Tupkaragan district. He was born in 1948 in Fort Shevchenko. He graduated from KazPTI in 1972. In 1972-2005, he worked on various positions: in Zhetybai, Embimunay quarry, Mangistau Ulutas Production Association, Mangyshlak auylkurylys-12 trust, Mangistaumunaykurylys trust, Ozen firm and others. Since 2006, he has been the director of Kubyr LLP. He is the recipient of badges like "40 years of Uzen oil field", "100 years of Kazakhstan oil".  \\ 
\midrule
Culinary and Food & \foreignlanguage{russian}{Қазақтың ұлттық тағамы төстікті дайындау қадамдарын сипаттап беріңіз.} & \foreignlanguage{russian}{Төстікті дайындау үшін алдымен қойдың төс етін құйқасымен бірге алып, екіге бөледі. Оны істікке шаншып, сексеуіл шоғына қақтайды. Кейін төстікті терісімен бірге үйітіп, тұздайды. Бірнеше күн бойы күн тигізбей кептіреді, тұз сіңгеннен кейін отқа қақтап пісіреді.} \\
 & Describe the preparation steps of the Kazakh national dish tostik (brisket). & To prepare tostik, first take the lamb brisket with the skin and cut it in half. Poke it with a skewer and fry it on saxaul coal. Then singe the brisket together with the skin and salt it. It has to dry for several days with no sun exposure, and after the salt is absorbed, cook it on fire.  \\ 
\midrule
Cultural Institutions & \foreignlanguage{russian}{Қазақстандағы Отырар руханияты мұражай-кітапханасына енген көне заман кітаптары қалай жасалғанын түсіндіріңіз.} & \foreignlanguage{russian}{Көне заман кітаптары қолмен қауырсын қаламмен жазылып, Тараз ешкісінің терісіне түсірілген. Терілер қаттап, түйеге тиеп тасымалданған, әріптер өшпей, жүздеген жылдарға жетіп тұрған.} \\
 & Explain how ancient books, which are included in the Museum-Library of Otrar spirituality in Kazakhstan, were created. & Ancient books were written by hand with a quill pen and drawn on the skin of a Taraz goat. When the skins got hard, they were loaded on camels, the letters were not erased, they stayed for hundreds of years.  \\ 
\midrule
Culture and Traditions & \foreignlanguage{russian}{Қазақ халқының зооморфтық ою-өрнектері мен олар бейнелейтін элементтерді сипаттаңыз.} & \foreignlanguage{russian}{Қазақтың зооморфтық ою-өрнектері жануарлар әлемін бейнелейді. Олардың ішіндегі ең кең тарағаны – мүйіздерді бейнелеу. Мүйіз тәрізді доғалы «қошқар мүйіз», «мүйіз», «қос мүйіз», «сынық мүйіз» тәрізді өрнектер кездеседі. Сонымен қатар, «түйе мойын», «құс қанаты», «қаз мойны» сияқты жануарлардың басқа да дене мүшелері бейнеленеді. Бұл өрнектер үй заттарын, жиһаз бен қару-жарақ бұйымдарын сәндейді.} \\
& Describe the zoomorphic ornaments of the Kazakh people and the elements they represent. & Kazakh zoomorphic ornaments represent the animal world. The most common of them is the depiction of horns. There are horn-shaped arched patterns such as "koshkar muyiz" (ram's horn), "muyiz" (horn), "kos muyiz" (double horn), "synyk muyiz" (broken horn). In addition, other body parts of animals are depicted, such as "tuye moiyn" (camel neck),"kus kanaty" (bird wing),"kaz moiny" (goose neck). These patterns adorn household items, furniture and weapons.  \\ 
\midrule
Education and Academia & \foreignlanguage{russian}{Қазақстан тілтанушы ғалымы Ысқақов Ахмеди Ысқақұлының қазақ тілін дамытудағы үлесін түсіндіріңіз.} & \foreignlanguage{russian}{Ысқақов Ахмеди Ысқақұлы қазақ тілтануының тарихы мен грамматикасы, лексика және оқыту методикасы саласындағы зерттеулерімен танымал болды. Ол «Қазақ тілінің сөздігі» мен «Абай шығармаларының сөздігі» сияқты шығармаларды құрастырып, «Абай тілі сөздігін» (1968) басқарды. 200-ден астам ғылыми еңбектің авторы.} \\
 & Explain the contribution of Kazakh linguist Akhmedi Iskakov to the development of the Kazakh language. & Iskakov Akhmedi Iskakuly became famous for his research in the field of history and grammar of Kazakh linguistics, vocabulary and teaching methods. He wrote such works as the "Dictionary of the Kazakh language" and the "Dictionary of Abai's works", and headed the writing of the "Dictionary of the Abai language" (1968). He is the author of more than 200 scientific papers.  \\ 
\midrule
Health and Medicine & \foreignlanguage{russian}{Қазақстандық хирург пен медицина ғылымдарының докторы Абдоллаев Шәдібай Нұрғазыұлының кәсіби жетістіктерін сипаттаңыз.} & \foreignlanguage{russian}{Абдоллаев Шәдібай Нұрғазыұлы хирург, медицина ғылымдарының докторы және профессор. Ол Қазақ КСР-інің еңбек сіңірген дәрігері атағын алды. Бұл ғалымның негізгі ғылыми-зерттеулері жедел хирургияның мәселелеріне бағытталған, әсіресе, ерте кезеңдегі релапаротомия және жүкті әйелдерге хирургиялық көмек көрсету.} \\
 & Describe the professional achievements of the Kazakh surgeon and Doctor of Medical Sciences Abdollayev Shadibay Nurgazyuly. & Abdollayev Shadibay Nurgazyuly is a surgeon, Doctor of Medical Sciences and professor. He received the title of honored doctor of the Kazakh SSR. The main research of this scientist is focused on the problems of emergency surgery, especially relaparotomy at an early stage and surgical care for pregnant women.  \\ 
\midrule
Historical Figures & \foreignlanguage{russian}{Қазақ күйші-композиторы Махамбет Бөкейхановтың Қазақстан музыкалық мәдениетіне қосқан үлесін түсіндіріңіз.} & \foreignlanguage{russian}{Махамбет Бөкейханов қазақ күйлерін жаздырушы, композитор. Ол Құрманғазы, Дәулеткерей, Әлікей сияқты күйшілердің мұрасын біздің заманға жеткізіп, өзі де күй шығарған. 1933 жылы студенттердің халық аспаптары оркестрін ұйымдастырып, ұлттық музыка аспаптарын жетілдіру үшін шеберханада жұмыс істеді. Оның еңбегі арқылы көптеген қазақ күйлері заманымызға жетті.} \\
 & Explain the contribution of the Kazakh kuishi-composer Makhambet Bukeikhanov to the musical culture of Kazakhstan. & Makhambet Bokeikhanov is a composer of Kazakh kuis. He brought the legacy of such kui performers as Kurmangazy, Dauletkerey, Alikey to our time and wrote kui himself. In 1933 he organized a student orchestra of folk instruments and worked in a workshop to improve national musical instruments. Thanks to his work, many Kazakh kuis have reached our time.  \\ 
\bottomrule
\end{tabular} }
\caption{Example of created instructions on \texttt{CultSet}.}
\label{tab:example_cultset}
\end{table*}

\begin{table*}[b!]
\scriptsize
\centering
\scalebox{0.8}{
\begin{tabular}{p{0.15\textwidth}p{0.2\textwidth}p{0.7\textwidth}}
\toprule
Category & Instruction & Output \\
\midrule
History and Archaeology & \foreignlanguage{russian}{Қазақ халқының Албан тайпасының тарихи мекендеген жерлерін сипаттаңыз.} & \foreignlanguage{russian}{Қазақ халқының Албан тайпасы негізінен Жетісу өңірінде, әсіресе қазіргі Алматы облысының Кеген мен Нарынқол аудандарында мекендеген. Олар Іле өзенінің сол жағында, Қырғыз шекарасына дейінгі аймақта және Верный мен Жаркент уездерінде шоғырланған. Бұл аймақ Іле Алатауының биік жоталарына дейінгі жерлерді қамтиды.} \\
\\ \\ & Describe the historical habitats of the Albanian tribe of the Kazakh people. & The Albanian tribe of the Kazakh people lived mainly in the Zhetysu region, especially in the Kegen and Narynkol districts of the modern Almaty region. They were concentrated on the left side of the Ili River, in the region up to the Kyrgyz border and in the Verny and Zharkent counties. This region covers lands up to the high ridges of the Trans-Ili Alatau.  \\ 
\midrule
Literature and Writing & \foreignlanguage{russian}{Қазақ ақыны Абайдың қазақ қоғамындағы теріс пиғылдарды қалай сынға алғанын сипаттаңыз.} & \foreignlanguage{russian}{Абай қазақ қоғамындағы әділетсіздік, парақорлық, арамзалық және еріншектік сияқты теріс қылықтарды ашық сынға алады. Ол халықтың бір-бірін аңдып, ұрлық жасап, шын келісімге жете алмайтынына назар аударады. Ақын өнер, әдебиет және ғылымға көңіл бөліп, қоғамдық тәрбиеге баса мән береді.} \\
\\ \\ & Describe how the Kazakh poet Abai criticized the negative aspects of Kazakh society. & Abai openly criticizes such wrongdoings in Kazakh society as injustice, bribery, dishonesty and laziness. He draws attention to the fact that the people follow each other, steal and do not reach a true agreement. The poet pays attention to art, literature and science, and pays special attention to public education.  \\ 
\midrule
Media and Communication & \foreignlanguage{russian}{Гүлмира Әлібекқызы Қаракөзованың қазақ журналистика саласындағы жетістіктерін түсіндіріңіз.} & \foreignlanguage{russian}{Гүлмира Қаракөзова - тәжірибелі журналист. Ол «Возвращенные имена» (1997) арнайы репортаждары топтамасының, «Паритет» (2002-2004) тележобасының және «Жұлдызды Астана» (2005 жылдан) радиожобасының авторы. 2004 жылдан бастап «Астана» радиосын басқарады. «Мәдениет қайраткері» құрмет белгісімен (2006) және «Астана» медалімен (2008) марапатталған.} \\
\\ \\ & Explain the achievements of Gulmira Alibekkyzy Karakozova in the field of Kazakh journalism. & Gulmira Karakozova is an experienced journalist. She is the author of a series of special reports "Returning names" (1997), the TV project "The Parity" (2002-2004) and the Radio Project "Starry Astana" (since 2005). Since 2004, she has been leading Astana radio. She was awarded the Badge of Honor "Madeniyet kairatkeri" (2006) and the medal "Astana" (2008).  \\ 
\midrule
Miscellaneous & \foreignlanguage{russian}{Қазақстан заңгері мен еңбек сіңірген прокуроры Қабдолла Досбалаұлы Төлегеновтың өміріндегі маңызды қызметтерді сипаттаңыз.} & \foreignlanguage{russian}{Қабдолла Досбалаұлы Төлегенов 1952 жылы туған. Ол 1975 жылы Ташкент мемлекеттік университетінің заң факультетін бітірген. Әр жылдарда түрлі қызметтерді атқарған, соның ішінде Шымкент, Алматы облыстарының прокуроры, ҚР Қауіпсіздік кеңесінің Хатшылығы меңгерушісінің орынбасары және басқалары. 2008 жылдан бері Шығыс Қазақстан облысының прокуроры. Ол «Құрмет» орденімен және басқа медальдармен марапатталған.} \\
\\ \\ & Describe the most important job positions in the life of the lawyer and honored prosecutor of Kazakhstan Kabdulla Dosbalauly Tulegenov. & Kabdolla Dosbalauly Tolegenov was born in 1952. He graduated from the faculty of law of Tashkent State University in 1975. Over the years, he held various positions, including prosecutor of Shymkent, Almaty regions, deputy head of the Secretariat of the Security Council of the Republic of Kazakhstan and others. He is the prosecutor of the East Kazakhstan region since 2008. He was awarded the Order "Kurmet" and other medals.  \\ 
\midrule
Nature and Environment & \foreignlanguage{russian}{Қазақстандағы Көлсай көлдерінің ұлттық табиғи паркінде кездесетін ерекше өсімдіктерді сипаттаңыз.} & \foreignlanguage{russian}{Көлсай көлдерінің ауданында ерекше сирек кездесетін өсімдіктерден өрік, Іле барбарисі, долана, Альберт құртқашашы бар. Аудан экологиялық тұрғыдан таза және жабайы өсімдіктерге бай. Бұл өсімдіктердің көбі дәрілік қасиетке ие.} \\
\\ \\ & Describe the unique plants found in the Kolsai Lakes National Natural Park in Kazakhstan. & In the area of Kolsai lakes, especially rare plants can be found - apricot, Ili barbaris, hawthorn, Iris albertii. The area is ecologically clean and rich in wild plants. Many of these plants have medicinal properties.  \\ 
\midrule
Politics & \foreignlanguage{russian}{Қазақстандық қоғам қайраткері Нұрлан Жылқышиевтің өмірбаяндық фактілерін суреттеңіз.} & \foreignlanguage{russian}{Нұрлан Жылқышиев 1953 жылы 3 мамырда Тараз қаласында туған. Ол Шымкент педагогикалық институтын және Алматы ауыл шаруашылығы институтын бітірген. 1998 жылы АҚШ-тың Тынық мұхиты жағалауы университетінде философия докторы дәрежесін қорғаған. Қабыланбек агрошарап компаниясының бас директоры болып қызмет атқарған және 2004 жылдан парламенттің мәжіліс депутаты. Ол әртүрлі марапаттарға ие болған, соның ішінде 'Тың және тыңайған жерлерді игерудің 50 жылдығы' төсбелгісі.} \\
 & Describe the biographical facts of the Kazakh public figure Nurlan Zhylkyshiev. & Nurlan Zhylkyshiev was born on May 3, 1953 in Taraz. He graduated from the Shymkent Pedagogical Institute and the Almaty Agricultural Institute. In 1998, he defended his doctorate in philosophy at the University of the Pacific coast of the United States. He worked as the head director of the agro-processing company Kabylanbek and since 2004 has been a deputy of the Mazhilis of the Parliament. He has received various awards, including the "50th anniversary of the development of Virgin and fallow lands" badge.  \\ 
\midrule
Sciences and Humanities & \foreignlanguage{russian}{Еділ Ерғожаұлы Ерғожинның химия саласында Қазақстанның мемлекеттік сыйлығын алғаны рас па?} & \foreignlanguage{russian}{Иә, Еділ Ерғожаұлы Ерғожин химия ғылымының докторы, профессор, Қазақстанның мемлекеттік сыйлығының иегері.} \\
\\ \\ & Is it true that Edil Yergozhauly Yerhozhin received the State prize of Kazakhstan in the field of chemistry? & Yes, Edil Yergozhauly Yergozhin is a doctor of chemical sciences, professor, winner of the State prize of Kazakhstan.  \\ 
\midrule
Sports and Recreation & \foreignlanguage{russian}{Қазақстандық пауэрлифтер-паралимпиадшы Давид Дегтярев 2020 жылғы жазғы Паралимпиада ойындарында қандай жетістіктерге жетті?} & \foreignlanguage{russian}{Қазақ пауэрлифтері Давид Дегтярев 2020 жылғы жазғы Паралимпиада ойындарында 54 кг дейінгі салмақ дәрежесінде жарысқа қатысып, 170 кг, 172 кг және 174 кг салмақтарды көтеру арқылы алтын медаль иегері атанды.} \\
 & What did Kazakhstani powerlifter-paralympian David Degtyarev achieve at the 2020 Summer Paralympic Games? & Kazakh powerlifter David Degtyarev competed at the 2020 Summer Paralympic Games in the weight category up to 54 kg and became the holder of a gold medal by lifting weights 170 kg, 172 kg and 174 kg.  \\ 
\midrule
Transportation and Infrastructure & \foreignlanguage{russian}{Қазақстандағы Ақтау Халықаралық Әуежайының тарихын қысқаша сипаттаңыз.} & \foreignlanguage{russian}{Ақтау Халықаралық Әуежайы 1983 жылы ашылған және көптеген ірі жүк тиейтін әуе кемелерін қабылдауға жабдықталған. 1996 жылы акционерлік қоғам ретінде құрылды. 2007 жылы жаңа жолаушылар терминалын салу мен басқаруға келісім жасалды. 2009 жылы жаңа терминал пайдалануға берілді, ол сағатына 450 жолаушыға қызмет көрсетеді. Әуежай жаңғыртулар нәтижесінде ТМД елдері арасында үздік әуежай атағын алды.} \\
\\ \\ & Briefly describe the history of Aktau International Airport in Kazakhstan. & Aktau International Airport was opened in 1983 and is equipped to receive many large cargo aircraft. It was established in 1996 as a joint stock company. In 2007, an agreement was signed for the construction and management of a new passenger terminal. In 2009, a new terminal was put into operation, serving 450 passengers per hour. As a result of the modernization, the airport received the title of the best airport among the CIS countries.  \\ 

\bottomrule
\end{tabular} }
\caption{Example of created instructions on \texttt{CultSet}.}
\end{table*}

\clearpage
\section{Inner-annotator agreement}
\label{app:inner-annot-appendix}
\subsection{Inner-Annotator Agreement for Preliminary Study}
\label{app:inner-annot-prelim}

\begin{figure*}[h!]
    \centering
    \begin{minipage}[b]{0.3\textwidth}
        \centering
        \includegraphics[scale=0.35]{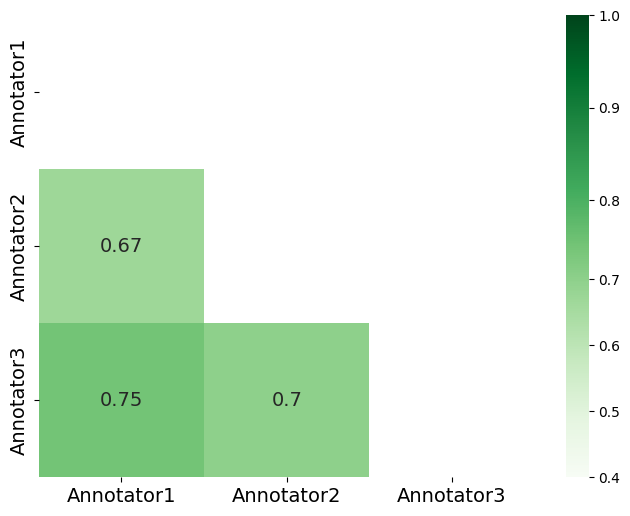}
    \end{minipage}
    \hfill
    \begin{minipage}[b]{0.3\textwidth}
        \centering
        \includegraphics[scale=0.35]{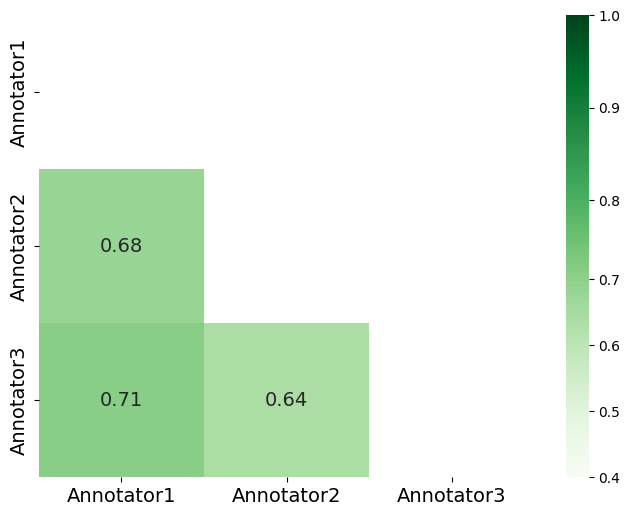}
    \end{minipage}
    \hfill
    \begin{minipage}[b]{0.3\textwidth}
        \centering
        \includegraphics[scale=0.35]{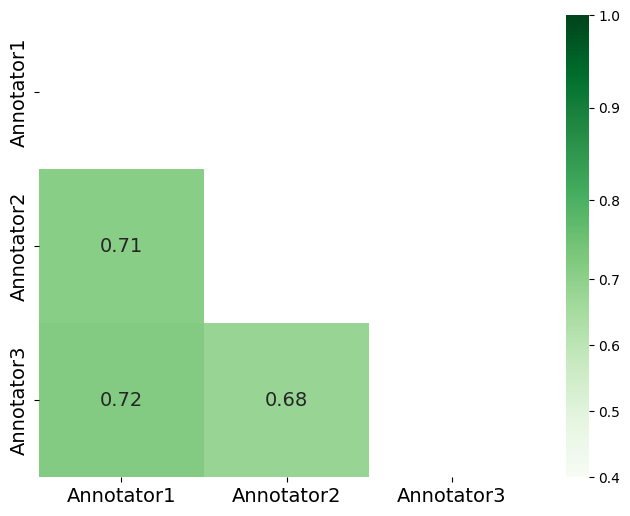} 
    \end{minipage}
    \caption{Inner annotator agreement across annotators for correctness, completeness, and fluency, measured using Pearson correlation.}

    \label{fig:inner-annotator-agreement}
\end{figure*}

\subsection{Inner-Annotator Agreement for Generation Evaluation}
\label{app:inner-annot-gen-eval}

\begin{figure}[h!]
    \centering
    \includegraphics[scale=0.45]{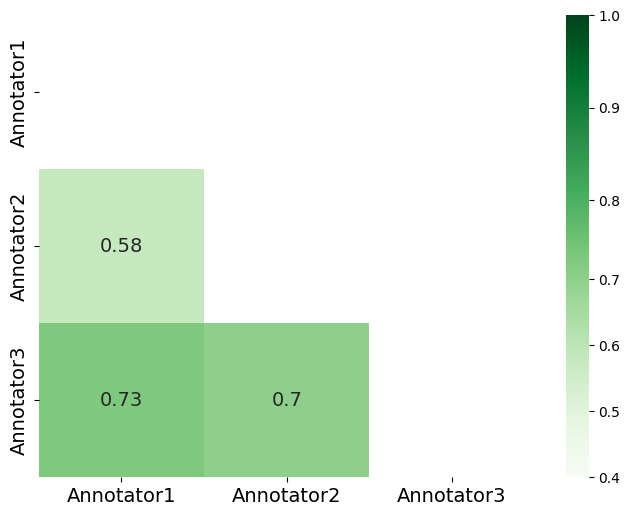} 
    \caption{Inner-annotator agreement for generation evaluation, measured using Cohen’s Kappa.}

    \label{fig:inner-annotator-generation}
\end{figure}

\clearpage

\section{MCQ Examples}
\label{sec:mcq_samples}
\begin{table*}[ht]
\centering
\renewcommand{\arraystretch}{1.5} 
\scriptsize 
\begin{tabular}{|p{2cm}|p{5cm}|p{5cm}|p{1cm}|}  
\hline
\textbf{Question Set}&\textbf{Kazakh} & \textbf{English translation} & \textbf{Correct Answer}\\ \hline
Dastur & \foreignlanguage{russian}{Тыйым деген не? \newline
A) Қазақ халқының той рәсімі \newline
B) Қазақ халқының тәрбиелік құралы \newline
C) Қазақ халқының музыкалық құралы \newline
D) Қазақ халқының аспап-құралы\newline
} & 
What is a tiyim? \newline
A) A Kazakh wedding ceremony \newline
B) A Kazakh educational tool \newline
C) A Kazakh musical instrument \newline
D) A Kazakh musical instrument\newline
& B \\
\hline
Dastur & \foreignlanguage{russian}{
Белкөтерер тағамы қандай адамдарға арнап дайындалады? \newline
A) Жас адамдарға \newline
B) Тісі болмаған, тамақ жей алмаған адамдарға \newline 
C) Спортшыларға
 \newline
D) Бала-шағаларға \newline
} & 
What kind of people is Belkoterer food prepared for? \newline
A) Young people \newline
B) People who have no teeth and cannot eat \newline 
C) Athletes
\newline
D) Children \newline
&
B \\
\hline
Constitution &
\foreignlanguage{russian}{
Конституция АДАМ ЖӘНЕ АЗАМАТ II бөлімінің  негізгі мәні қандай? \newline

A) Адамдың қадір-қасиетін қорлау \newline
B) Адамдың қадір-қасиетіне қол сұғу\newline
C) Адамдың қадір-қасиетіне қол сұғылмайды \newline
D) Адамдың қадір-қасиетін жою\newline
}&
What is the main idea of the Constitution's Section II, chapter on rights of people and citizens? \newline

A) Insulting human dignity \newline
B) Infringement on human dignity\newline
C) Human dignity is inviolable \newline
D) Destruction of human dignity\newline
&
C \\
\hline
Constitution &
\foreignlanguage{russian}{
Қазақстан Республикасының басқару нысаны қандай? \newline
A) Парламенттік басқару \newline
B) Президенттік басқару\newline
C) Монархиялық басқару \newline
D) Федерациялық басқару \newline
} & 
What is the form of government of the Republic of Kazakhstan? \newline
A) Parliamentary government \newline
B) Presidential government\newline
C) Monarchical government \newline
D) Federal government \newline
& B \\
\hline
Human Rights and Society &
\foreignlanguage{russian}{
ҚР Парламент Мәжілісі депутаты болу үшін қойылатын талаптарды атаңыз \newline
A) он жыл еңбек өтілі\newline
B) елу жастан аспау\newline
C) жоғары білім\newline
D) жиырма беске толу\newline

} & 
What are the requirements for becoming a deputy of the Mazhilis of the Parliament of the Republic of Kazakhstan?\newline
A) ten years of work experience\newline
B) not older than fifty years\newline
C) higher education\newline
D) twenty-five years of age\newline
& D\\
\hline
Human Rights and Society & 
\foreignlanguage{russian}{
Он алты жастан он сегіз жасқа дейінгі қызметкерлер үшін жұмыс уақытының ұзақтығы аптасына аспауы тиіс:\newline
A) 36 сағаттан\newline
B) 32 сағаттан\newline
C) 24 сағаттан\newline
D) 34 сағаттан\newline
}&

For employees aged sixteen to eighteen, the working hours per week shall not exceed:\newline
A) 36 hours\newline
B) 32 hours\newline
C) 24 hours\newline
D) 34 hours\newline&
A \\ \hline
\end{tabular}
\caption{Examples of MCQ.}
\label{tab:mcq_samples}
\end{table*}
\clearpage

\section{Example of conversational data}
\label{sec:conversational-data-sample}
\begin{table*}[ht]
\centering
\renewcommand{\arraystretch}{1.5} 
\scriptsize 
\begin{tabular}{p{6cm}|p{6cm}}  
\hline
\textbf{Kazakh} & \textbf{English translation} \\ \hline
\textcolor{darkgreen}{\textbf{\textit{Person 1 says:}}} \foreignlanguage{russian}{Қазақстанда жылжымайтын мүлік алу және сату кезінде заңдық талаптар маңызды рөл атқарады. Бұл процесс кейде ұзаққа созылуы мүмкін, әсіресе, меншік иелері бірнешеу болған жағдайда.} 
& \textcolor{darkgreen}{\textbf{\textit{Person 1 says:}}} In Kazakhstan, legal requirements play an important role when buying and selling real estate. This process can sometimes be prolonged, especially when there are multiple property owners. \\[0.3cm]

\textcolor{orange}{\textbf{\textit{Person 2 says:}}} \foreignlanguage{russian}{Иә, әсіресе, мүлікті мұрагерлік жолмен алған жағдайда немесе кәмелетке толмаған бала иелік еткенде қосымша құжаттар талап етіледі.} 
& \textcolor{orange}{\textbf{\textit{Person 2 says:}}} Yes, especially when the property is inherited or owned by a minor, additional documents are required. \\[0.3cm]

\textcolor{darkgreen}{\textbf{\textit{Person 1 says:}}} \foreignlanguage{russian}{Қазақстандағы жылжымайтын мүліктің сатып алу-сату үрдісі өте маңызды болып табылады. Сіз білесіз бе, кәмелетке толмаған баланың мүлкін сату үшін қандай құжаттар қажет?} 
& \textcolor{darkgreen}{\textbf{\textit{Person 1 says:}}} The process of buying and selling real estate in Kazakhstan is very important. Do you know what documents are required to sell the property of a minor? \\[0.3cm]

\textcolor{orange}{\textbf{\textit{Person 2 says:}}} \foreignlanguage{russian}{...} 
& \textcolor{orange}{\textbf{\textit{Person 2 says:}}} ... \\[0.3cm]

\textcolor{red}{\textbf{\textit{Gemma's response:}}} \foreignlanguage{russian}{"...Кәмелетке толмаған баланың мүлкін сату үшін ата-анасының немесе олардың орынбасарының қол қоюымен жасалған өтініш, баланың туу туралы куәлігі, мүліктің сатып алу-сату туралы құжаттар, мүліктің құқықтық мәртебесін анықтайтын құжаттар қажет."}  
& \textcolor{red}{\textbf{\textit{Gemma's response:}}} "...To sell the property of a minor, an application signed by the parents or their legal representatives, the child's birth certificate, documents regarding the purchase and sale of the property, and documents confirming the legal status of the property are required." \\ 

\hline
\end{tabular}
\caption{Example of conversational data on \texttt{GovSet}.}
\label{tab:conversational-data-sample}
\end{table*}

\end{document}